%% file: main.tex
\documentclass[10pt,twocolumn,letterpaper]{article}

\usepackage{cvpr}      

\usepackage{fontawesome}
\usepackage{multirow}
\usepackage{arydshln}
\usepackage{comment}
\usepackage{subcaption}
\usepackage{booktabs}
\usepackage{cancel}
\usepackage{pifont}
\DeclareMathSizes{10}{9}{6}{5}
\definecolor{cvprblue}{rgb}{0.007,0.30,0.50}
\usepackage[pagebackref,breaklinks,colorlinks,allcolors=cvprblue]{hyperref}
\renewcommand{\paragraph}[1]{\noindent\textbf{#1}~}
\def\confName{CVPR}
\def\confYear{2025}
\usepackage{xcolor}
\definecolor{red}{RGB}{166, 25, 46}  
\definecolor{blue}{RGB}{2, 77, 127} 
\definecolor{green}{RGB}{43, 127, 62}
\title{VITAL: Vision-Encoder-centered Pre-training for LMMs \\ in Visual Quality Assessment}

\author{Ziheng Jia\textsuperscript{1*}, Linhan Cao\textsuperscript{1*}, Jinliang Han$^1$, Zicheng Zhang$^2$, Jiaying Qian$^1$, Jiarui Wang$^1$, Zijian Chen$^1$\\
Guangtao Zhai$^{1,2}$, Xiongkuo Min$^{1\diamondsuit}$\\
$^1$Shanghai Jiaotong University\\$^2$Shanghai Artificial Intelligence Laboratory
}

\begin{document}
\maketitle
\input{sec/0_abstract}  
\begin{figure}
    \centering
    \includegraphics[width=0.98\linewidth]{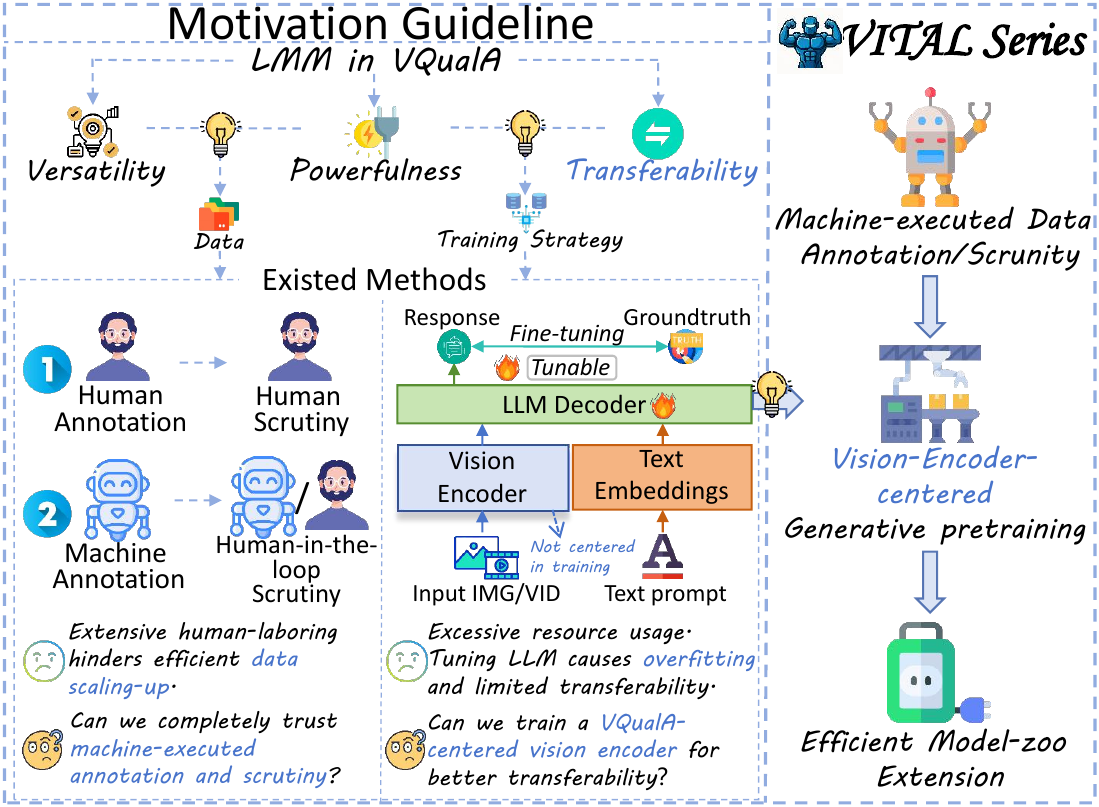}
    \vspace{-7pt}
    \caption{Comprehensive \textbf{dataset} construction underpins the versatility and performance of VQualA LMMs, while an effective \textbf{training strategy} is also essential for improving transferability and scalability.
Most existing works focus on a single visual modality or task and depend heavily on human annotation, limiting scalability and versatility. In addition, fine-tuning the LLM decoder often causes overfitting, reducing transferability. These limitations serve as our motivation.
}
   
    \label{fig:teaser}
    
\end{figure}
\input{sec/1_intro}
\begin{figure*}
    \centering
    \includegraphics[width=0.99\linewidth]{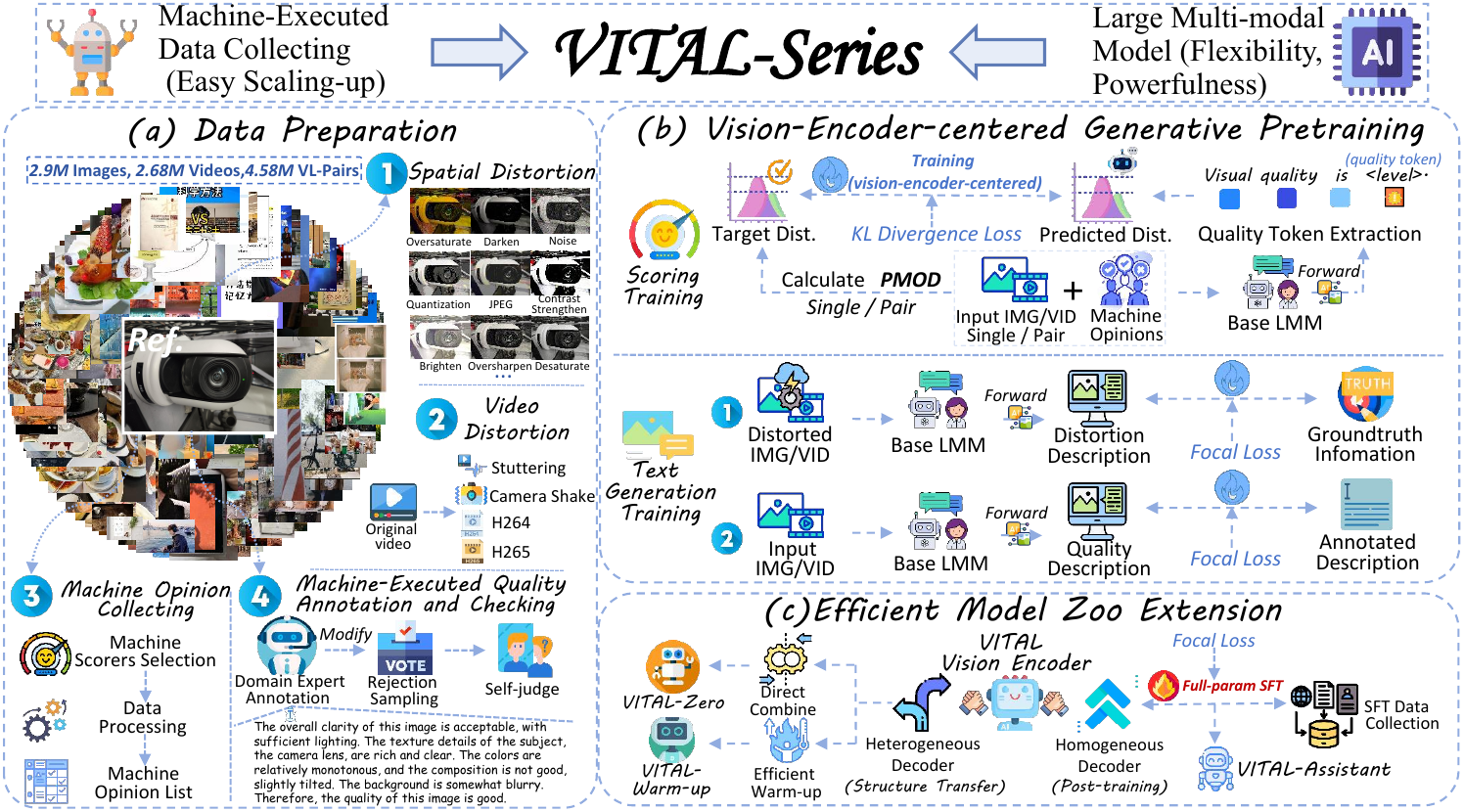}
     \vspace{-12pt}
    \caption{Overall workflow of the dataset preparation and the pretraining process of our VITAL Series.}
    \label{fig:workflow}
   
\end{figure*}
\input{sec/2_related_works}
\begin{figure*}
    \centering
    \includegraphics[width=0.98\linewidth]{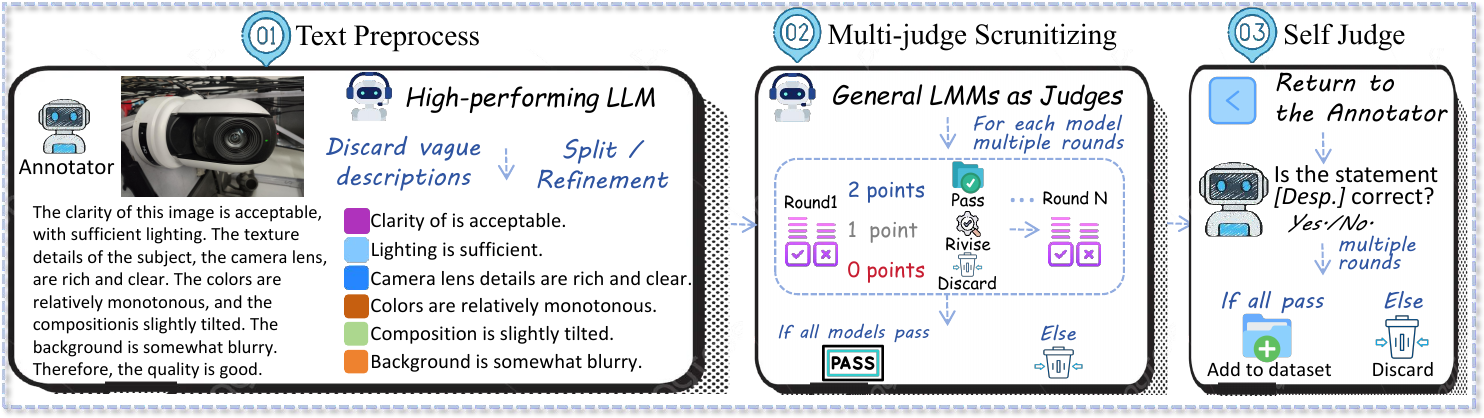}
     \vspace{-10pt}
    \caption{Machine-executed data annotation and quality scunity for the quality interpreting subtask. }
    \label{fig:annotation}
   
\end{figure*}
\input{sec/3_internvquala}

\input{sec/4_experiments}
\input{sec/5_conclusion}

\input{sec/X_suppl}

\clearpage
\onecolumn
\twocolumn
{
    \small
    \bibliographystyle{ieeenat_fullname}
    \bibliography{main}
}


\end{document}

%% file: sec/0_abstract.tex
\begin{abstract}
Developing a robust visual quality assessment (VQualA) large multi-modal model (LMM) requires achieving \textbf{versatility}, \textbf{powerfulness}, and \textbf{transferability}.
 However, existing VQualA LMMs typically focus on a single task and rely on full-parameter fine-tuning, which makes them prone to overfitting on specific modalities or task types, thereby limiting their generalization capacity and transferability. To address this, we propose a \textbf{vision-encoder-centered generative pre-training} pipeline and develop the \textbf{VITAL-Series} LMMs.
(1) We adopt a machine-executed annotation–scrutiny paradigm, constructing over \textbf{$4.5M$} vision–language (VL) pairs—the \textbf{largest VQualA training dataset to date}. (2) We employ a multi-task training workflow that simultaneously enhances the model’s quantitative scoring precision and strengthens its capability for quality interpretation across both image and video modalities.  (3) Building upon the vision encoder, we realize an \textbf{efficient model zoo extension}: the model zoo exhibits strong zero-shot performance, and each paired decoder requires only a swift warm-up using less than $1/1000$ of the pre-training data to achieve performance comparable to the fully trained counterpart. Overall, our work lays a cornerstone for advancing toward the \textbf{foundation LMM for VQualA}.

\end{abstract}

%% file: sec/1_intro.tex
\section{Introduction}
With the rapid advancement of large multi-modal models (LMMs)~\cite{achiam2023gpt,gao2023llama,zhang2024video,li2024llava,bai2025qwen2,xu2025qwen3}, the field of  computer vision has undergone a transformative evolution. 
The strategy of \textbf{visual-language instruction tuning}~\cite{liu2023visual}, which harnesses extensive multi-modal instruction databases (MIDBs), has substantially improved the capabilities of LMMs in complex tasks~\cite{zhou2024mlvu,li2024mvbench,li2023seed,wang2024internvideo2,AIBench,zhang2025large}. In parallel, leveraging this to facilitate low-level perception-based vision tasks also shows significant promise. A typical subfield of this is perceptual visual quality assessment (VQualA).

Fundamentally, the development of a robust \textbf{LMM for VQualA} depends on the attainment of \textbf{versatility}—the ability to process commonly encountered visual modalities and to adapt to a range of tasks (including scoring (aligning with human mean opinion scores (MOSs)) and textual interpreting). Meanwhile, the model should demonstrate \textbf{powerfulness}, achieving high performance across multiple scenarios.
Furthermore, as different application scenarios and hardware affordances necessitate models with varying parameter sizes, it is equally important—though frequently overlooked—to ensure \textbf{transferability}: the potential to efficiently perform structural adaptation and model zoo extension based on the pretrained model.



\textbf{Training data} constitute the foundation that fuels the development of versatile and powerful LMMs in VQualA. From the functional perspective, the diversity of training data determines the capability boundary of the LMM. 
However, existing works are typically trained on datasets focused on a single modality or task, resulting in a restricted capability across multiple scenarios.
In terms of performance, the development of a high-performing VQualA LMM is also contingent upon the large-scale, high-quality datasets.
 However, the reliance on labor-intensive and costly human annotation or verification processes imposes significant constraints on data scalability.
Consequently, current VQualA MIDBs remain limited in scale and diversity, hindering their applicability. These observations lead to a critical question: \textit{Is it feasible to construct large-scale and task-comprehensive datasets with satisfactory quality solely through machine-executed annotation?}
We believe the answer is affirmative: (1) Recent LMMs demonstrate remarkably low-level perception abilities and exhibit reduced bias to external uncertainties compared to human annotation. (2) Machine-generated annotations can still exhibit enhanced robustness when supported by effective formatting strategies and weak-to-strong generalization mechanisms.

Moreover, a \textbf{well-designed training strategy} is also essential to achieving both powerfulness and transferability. 
Existing LMMs for VQualA, often trained via full-parameter (with the large language model (LLM) decoder tuned) supervised fine-tuning (SFT) on a fixed structure, tend to overfit on specific data or task types, resulting in poor generalization performance. Through extensive analysis (detailed in the supplementary materials (\textit{Supp.}) \textit{Sec.~\ref{Supplementary Experiments}}), we identify the vision encoder as the core component of VQualA-centric LMMs. Meanwhile, the pre-training process~\cite{radford2018improving,radford2021learning} has been shown to facilitate inter-domain and cross-structural transferability.
Consequently, we argue that adopting a vision-encoder-centered pre-training paradigm constitutes a key solution.

The above discussions (summarized in Fig.~\ref{fig:teaser}) motivate our work. We introduce the \textbf{\underline{Vi}sion-Encoder-centered Pre-\underline{t}r\underline{a}ining for \underline{L}MMs in VQualA (VITAL)}, a vital training framework built entirely upon machine-annotation. Our main contributions are in threefold:

\begin{itemize}
    \item We construct a machine-annotated large-scale pre-training dataset comprising $4.58M$ vision-language (VL) pairs. The dataset covers two major tasks: \textbf{visual quality scoring} and \textbf{text generation}. We also design detailed data processing procedures to maximize the reliability of these machine-generated labels.
    
    \item We conduct the \textbf{vision-encoder-centered generative pre-training} on the tasks. For quality scoring, we employ the \textbf{proxy machine opinion distribution (PMOD) prediction} strategy to mitigate potential biases introduced by machine-produced weak labels. For text generation, we modify the loss function to dynamic \textit{focal loss}, which balances the learning rates of simpler and more complex targets while alleviating structure overfitting. The pre-training procedure produces the \textbf{VITAL Vision Encoder}.

    \item  This vision encoder facilitates \textbf{efficient model structural transfer}, enabling the construction of a scalable model zoo. The model zoo demonstrates strong zero-shot capabilities and, following post-training or efficient few-shot warm-up (with only $4000$ samples), achieves superior performance and generalization across diverse VQualA tasks and datasets, against state-of-the-art (SOTA) baselines.
\end{itemize}

\label{sec:intro}

%% file: sec/2_related_works.tex
\section{Related Works}
Fundamental visual quality assessment can be categorized into Image Quality Assessment (IQA)~\cite{zhai2020perceptual} and Video Quality Assessment (VQA)~\cite{min2024perceptual}.  IQA can be classified by content type, such as in-the-wild IQA~\cite{hosu2020koniq}, AI-generated content (AIGC) IQA~\cite{li2023agiqa}, panoramic IQA~\cite{sun2019mc360iqa}, among others.
 VQA extends IQA by incorporating temporal factors, covering various types as well, such as user-generated content (UGC)~\cite{tu2021ugc}, professionally generated content (PGC)~\cite{cheng2020screen,jia2024dsa,lu2022deep,yang2022blind}, and AI generated videos~\cite{zhang2024human,zhang2025q,chen2024gaia}. 
\begin{table}[!t]\tiny
    \centering
\renewcommand\arraystretch{0.9}
\renewcommand\tabcolsep{6pt}
\belowrulesep=0pt\aboverulesep=0pt
    \caption{Statistic summary of our pre-training dataset. Note that the total counts of images and videos include the overlapping content. Each \textit{pairwise-training} sample represents a single VL pair.}
    \vspace{-10pt}
   \resizebox{0.97\linewidth}{!}{\begin{tabular}{c|c|c|c}
    \hline
    \textbf{Modality} &\textbf{Subtask Types} &\textbf{\# Original~/~Sampled } & \textbf{\# V-L Pairs}  \\ \hline
    \multirow{4}{*}{\textit{\textbf{Image}}}&\multirow{1}{*}{\textit{Distribution}}& 5,000,000~/~1,000,000 &\multirow{1}{*}{1,000,000} \\
     &\multirow{1}{*}{\textit{Pair}}& 5,000,000~/~1,000,000 & 500,000 \\ 
     &\multirow{1}{*}{\textit{Distortion}}& 5,000,000~/~500,000 &500,000 \\
     &\multirow{1}{*}{\textit{Depict}}& 5,000,000~/~400,000 &400,000 \\ \hline
     \multirow{4}{*}{\textit{\textbf{Video}}}&\multirow{1}{*}{\textit{Distribution}}& 4,000,000~/~845,000 &\multirow{1}{*}{845,000} \\
     &\multirow{1}{*}{\textit{Pair}}& 4,000,000~/~1,000,000 & 500,000 \\ 
     &\multirow{1}{*}{\textit{Distortion}}& 4,000,000~/~435,144 &  435,144 \\
     &\multirow{1}{*}{\textit{Depict}}& 4,000,000~/~400,000 & 400,000\\ \hline
     \multicolumn{1}{c|}{\textbf{Overall}}& \multicolumn{2}{c|}{\# \textit{Images}: 2,900,000 \# \textit{Videos}: 2,680,144} &\multirow{1}{*}{4,580,144}\\
     \hline
\end{tabular}
 }
\vspace{-8pt}
\label{tab:dataStatistic}
\end{table}

Studies have already leveraged LMMs to address specific VQualA tasks. \textit{Q-Align}~\cite{wu2024q1} lays the foundation of the LMM-based quality scoring using the one-hot probability estimation strategy.  \textit{Compare2Score}~\cite{zhu2024adaptive} tackles subjective label scarcity using pairwise preference as supplementary labels, and \textit{DeQA-Score}~\cite{you2025teaching} further explores leveraging soft-labeled MOS distribution to boost IQA scoring. \textit{Q-Instruct}~\cite{wu2024q}, \textit{Aes-Expert}~\cite{huang2024aesexpert}, and \textit{VQA\textsuperscript{2}}~\cite{jia2024vqa} pioneer in training LMMs with quality interpreting capabilities in image technical quality, image aesthetics, and video quality, respectively. \textit{Co-Instruct}~\cite{wu2024towards} and \textit{DepictQA}~\cite{you2024depicting,you2024descriptive} focus on image pair quality comparison and joint analysis tasks. However, as mentioned above, these LMMs are limited to single modalities or tasks and lack a transferable strategy for efficiently extending the model zoo. 

Diverse VQualA MIDBs have also been constructed for instruction tuning. These include \textit{Q-Align-DB}~\cite{wu2024q1} and \textit{VQA\textsuperscript{2}-Stage2}~\cite{jia2024vqa}, which are reformulated from subjective labels used for quality scoring. Additionally, fully human-annotated MIDBs, such as \textit{Q-Pathway-200K}~\cite{wu2024q}, \textit{AesMMIT}~\cite{huang2024aesexpert}, and \textit{VQA\textsuperscript{2}-Stage3}~\cite{jia2024vqa} exemplify human annotation and revision approaches to dataset construction.  Other datasets, like \textit{OmniVQA-Chat-400K}~\cite{jia2025scaling} and \textit{DepictQA-495K}~\cite{you2024depicting}, are based on machine-generated annotations/scutiny with human-in-the-loop. Nevertheless, the construction of these MIDBs still relies on manual labor, which prevents rapid data scaling-up.  All the above issues also motivate our development of the \textit{VITAL Series}.

%% file: sec/3_internvquala.tex
\section{The VITAL Series}
   
\subsection{Data Preparation}
Our dataset is selected from a huge candidate pool containing $5M$ images and $4M$ videos. It consists of two major task types: visual quality scoring and text generation (with distortion recognition and quality interpretation subtasks), comprising $4.58M$ VL pairs. Statistics are shown in Tab.~\ref{tab:dataStatistic}. The data preparation pipeline is provided in Fig.~\ref{fig:workflow}.

\paragraph{Candidate Pool Selection.}
Considering both the content diversity and the scalability of data collection, we advocate sourcing data from in-the-wild content.
We employ a minimally constrained strategy for candidate pool selection. Specifically, we retain only the basic controls—such as video duration ($1$-$20$s) and image resolution (no lower than $360\times480$ and no higher than $4K$)—to ensure compliance with training efficiency. 
Video data are randomly sampled from the \textit{Panda70M}~\cite{chen2024panda}, while image data are crawled from online platforms (methods detailed in \textit{Supp. Sec.~\ref {Dataset Construction}}). 


\paragraph{Quality Scoring Task.} Earlier quality scoring datasets typically rely on rigorous subjective experiments. Multiple human experts are required to rate the designated visual signals' quality under strictly controlled experimental environments and standardized protocols, thereby producing the opinion scores~\cite{international2023methodologies}. We propose that machine annotators can substitute the procedures of human scoring, since the differences in model design and architecture among various machine annotators represent distinct \textbf{perception perspectives}, effectively mirroring human individuality and diversity.
Meanwhile, although the annotations produced by individual models are generally less accurate than humans, aggregating multiple machine labels into an opinion distribution provides a more robust representation. This stems from the fact that the uncertainty of machine annotations is inherently reflected in the aggregated distribution, and estimating uncertainty can significantly enhance the LMM robustness~\cite{sun2025probabilistic}.
Specifically, we select six no-reference models that have demonstrated strong performance in objective VQA and IQA tasks, respectively\footnote{VQA: \textit{Minimalistic-VQA-VII}~\cite{sun2024analysis}, \textit{Minimalistic-VQA-IX}, \textit{FAST-VQA}~\cite{wu2023neighbourhood},   \textit{DOVER}~\cite{wu2023exploring}, \textit{Q-Align}~\cite{wu2024q1}, \textit{KVQ}~\cite{qu2025kvq}. IQA: \textit{TOPIQ-NR}~\cite{topiq}, \textit{TReS}~\cite{golestaneh2022no}, \textit{LIQE}~\cite{zhang2023blind}, \textit{ARNIQA}~\cite{agnolucci2024arniqa}, \textit{QualiCLIP}~\cite{agnolucci2024quality}, \textit{Q-Align}.}. To ensure consistency, all VQA models are pretrained on the \textit{LSVQ (train)} \cite{ying2021patch}, with the IQA ones pretrained on the \textit{KonIQ-10K (train)} \cite{hosu2020koniq}. The output scores of each model are all mapped to  $[0,1]$ (with videos and images separated, detailed in \textit{Supp. Sec.~\ref{PMOD Construction}})  and discretized into $5$ quality levels (\textit{high}, \textit{good}, \textit{fair}, \textit{poor}, and \textit{low}) with an interval of $0.2$. 



\paragraph{Text Generation Task.}
This task is designed to enhance the LMM’s \textbf{sensitivity to visual quality cues} in text generation. 
Since human VQualA primarily focuses on \textbf{visual distortions}, the most perceptually decisive factors, we prioritize this aspect in our dataset construction.
Specifically, we select $25$ types of spatial distortions (following the methodology in \textit{KADIS-700K}~\cite{lin2019kadid}) and four types of video-specific distortions. For spatial distortions, we define five severity levels: \textit{mild}, \textit{noticeable}, \textit{relatively severe}, \textit{severe}, and \textit{very severe}. For video distortions, we set three levels: \textit{mild}, \textit{noticeable}, and \textit{severe}.  For each image or video in the candidate pool that satisfies the selection criteria, we randomly assign a distortion type and a severity level, which are then applied uniformly across the entire extent of the sample (detailed in \textit{Supp. Sec.~\ref{Dataset Construction}}).
Each distortion is recorded in text format as ``\textit{[severity level]–[distortion type]}”, which is then refined to form the training-ready VL pair. 

In addition to distortion recognition, the capacity to deliver a textual description of visual quality attributes is equally essential. We employ a \textbf{rejection-sampling} approach characterized by ``domain-expert annotation with general LMM judgment” in the machine annotation stage. 
Specifically, we utilize the \textit{VQA\textsuperscript{2}-Assistant-Enhanced}~\cite{jia2024vqa} to generate comprehensive visual quality annotations for each sample. Subsequently, a proprietary LLM (we use the \textit{GPT-4o-mini}) is employed to refine and paraphrase these annotations. All vague statements (e.g., ``the quality of this image/video is good”) are removed during this process. The yielded descriptive paragraph is then decomposed into concise, sentence-level statements, each paired with the corresponding visual sample to form a VL pair. For each pair, we prompt high-performing LMMs (\textit{GPT-5-250807}, \textit{Gemini-2.5-Flash}~\cite{mallick2025gemini}, and \textit{Qwen-VL-Max-250813}~\cite{bai2023qwen}) to serve as the ``judges" and independently vote on their correctness for three rounds. If all judges completely agree with the match of the statement and the visual signal in all rounds, the pair is passed. Conversely, if any of the judges identifies a substantial deviation in any round, the corresponding VL pair is discarded. In cases where a judge reports minor inaccuracies or partial inconsistencies between the statement and the visual content, we adopt its revision suggestions. 

Inspired by \cite{wang2025llava}, we find that the annotator itself can also serve as an effective self-critical judge. Therefore, in the next step, we adopt a ``\textbf{self-judge}” strategy for further data filtering. All accepted statements in the former procedure are re-evaluated by the annotator using semantically equivalent but structurally varied prompts. Each statement undergoes three rounds of consecutive binary judgments; only those receiving consistent positive responses are retained and included in the dataset, while all others are discarded (the annotation and scrutiny process is visualized in Fig.~\ref{fig:annotation} and related prompts are all provided in \textit{Supp. Sec.~\ref{prompt}}).

\subsection{Generative Pre-training}
The \textbf{vision-to-text generative pre-training} paradigm~\cite{ke2023vila,liu2023visual} is a commonly adopted LMM pre-training strategy. Given that full-parameter fine-tuning is computationally expensive and often leads to overfitting, it diminishes the model’s transferability. Hence, we center the training process on the vision encoder by \textbf{freezing the LLM component and all projectors} (the workflow is shown in Fig.~\ref{fig:workflow}). 
We adopt the architecture in \textit{VQA\textsuperscript{2}}~\cite{jia2024vqa}, wherein the \textbf{vision encoder} comprises the \textbf{image encoder} and the \textbf{motion extractor} (\textit{SlowFast-R50}~\cite{feichtenhofer2019slowfast}). The extracted image and motion feature tokens are concatenated to form the input visual token sequence. We utilize \textit{InternVL-3-8B-Instruct}~\cite{zhu2025internvl3} as the base model for the image encoder (\textit{InternViT-300M-448px}), the vision projector (\textit{MLP}), and the LLM component (\textit{Qwen2.5-7B}~\cite{team2024qwen2}). The specific details of the model architecture are presented in \textit{Supp. Sec.~\ref{structure}}.
During image-only input cases, the \textit{SlowFast} component is deactivated. 



\paragraph{Quality Scoring Training.} 
Inspired by and following \textit{DeQA-Score}~\cite{you2025teaching}, we adopt the \textit{PMOD prediction}—an effective way to leverage machine-generated scores as a form of weak supervision.
Specifically, for each visual input $V$, we first compute the mean $\mu$ and standard deviation $\sigma$  of the corresponding machine opinion list. Using these parameters, we initialize a \textit{Gaussian} distribution $\mathcal{N}\left(\mu, \sigma^2\right)$ to represent the sample’s \textit{PMOD}. We then perform linear adjustment to adapt the probability mass within the $5$ predefined quality intervals (\textit{high} ($0.8$-$1.0$), \textit{good} ($0.6$–$0.8$), \textit{fair} ($0.4$–$0.6$), \textit{poor} ($0.2$–$0.4$), and \textit{low} ($0$–$0.2$)), ensuring that the adjusted probabilities $p_i, i \in \{0,1,2,3,4\}$ sum to $1$ and preserve the same mean value. 
During training, we adopt the commonly used target template ``\textit{The quality of this image/video is \textit{[level]}}". For all tokens preceding the \textit{[level]} token (at index $i_\textit{level}$), we apply the \textit{cross-entropy (CE)} loss to supervise text generation. For the \textit{[level]} token, the model outputs five logits corresponding to the predefined quality levels. These logits are converted into the predicted probabilities $p_i^{\textit{pred}}$ using the softmax processing. The \textit{Kullback–Leibler (KL) divergence} between the predicted and target \textit{PMOD} ($\mathcal{L}_\textit{kl}=\sum_{i=0}^{4} p_i \log (p_i / p_i^{\textit{pred}})$) is then employed.
Finally, the loss of the single visual signal input is designed as the weighted sum of the \textit{CE} and the \textit{KL} losses:
\begin{equation}
\label{single}
\mathcal{L_\textit{Scoring-single}} = - \frac{1}{L}\left(\gamma\sum_{\ell=0}^{i_\textit{level}-1} \log p(\mathbf{z_{\ell}} | \mathbf{Z_{\ell}})-\mathcal{L}_\textit{kl}\right),
\end{equation}
where $\mathbf{z_{\ell}}$ represents the ${\ell}$-th target label, $\mathbf{Z_{\ell}}$ denotes the input and target token sequence before $\mathbf{z_{\ell}}$, and $L$ represents the length of the target token sequence.
Emperically, $i_\textit{level}=L-1$ and $\gamma$ is set to $0.01$.

In subjective experiments, pairwise preferences are generally easier to obtain than absolute MOSs and are less susceptible to criteria and content biases. Analogously, training on \textit{pairwise preference PMOD} prediction can improve the LMM's generalization capability. 
Specifically, following \textit{DeQA-Score}~\cite{you2025teaching} and \textit{Thurstone Model}~\cite{thurstone2017law}, we treat each visual input’s \textit{PMOD} in a pair as independent. Consequently, the preference relationship between a visual signal pair 
$\left(V_{\textit{\uppercase\expandafter{\romannumeral1}}} \sim \mathcal{N}\left(\mu_{\textit{\uppercase\expandafter{\romannumeral1}}}, \sigma_{\textit{\uppercase\expandafter{\romannumeral1}}}^2\right),
V_{\textit{\uppercase\expandafter{\romannumeral2}}} \sim \mathcal{N}\left(\mu_{\textit{\uppercase\expandafter{\romannumeral2}}}, \sigma_{\textit{\uppercase\expandafter{\romannumeral2}}}^2\right)\right)$
can be modeled as the difference between two independent \textit{Gaussian} variables $\mathcal{N}\left(\mu_\textit{\uppercase\expandafter{\romannumeral1}}-\mu_\textit{\uppercase\expandafter{\romannumeral2}}, \sigma_\textit{\uppercase\expandafter{\romannumeral1}}^2+\sigma_\textit{\uppercase\expandafter{\romannumeral2}}^2\right)$. 
Consequently the \textit{pairwise PMOD} of the $V_{\textit{\uppercase\expandafter{\romannumeral1}}}$
surpassing $V_{\textit{\uppercase\expandafter{\romannumeral2}}}$ in visual quality can be modeled as: 
\vspace{-5pt}
\begin{equation}
  p^{\textit{pred}}(\textit{\uppercase\expandafter{\romannumeral1}}>\textit{\uppercase\expandafter{\romannumeral2}})=\Phi\left(\frac{\mu_\textit{\uppercase\expandafter{\romannumeral1}}^{\textit{pred}}-\mu_\textit{\uppercase\expandafter{\romannumeral2}}^{\textit{pred}}}{\sqrt{\left(\sigma_\textit{\uppercase\expandafter{\romannumeral1}}^{\textit{pred}}\right)^2+\left(\sigma_\textit{\uppercase\expandafter{\romannumeral2}}^{\textit{pred}}\right)^2}}\right),  
\end{equation}
where $\Phi\left(\dots\right)$ represents the standard \textit{Gaussian}.
To stabilize training, we manually set all the $\sigma^{\textit{pred}}$ as $1$. As the model outputs $5$ predicted quality level logits at index $i_\textit{level}$, the discrete \textit{pairwise PMOD} prediction can be derived from this.



When constructing the label for such preference pairs, we also account for \textbf{tie cases} in which the two visual signals exhibit similar quality (we set the ratio of \textit{better}, \textit{worse}, and \textit{tie} samples to $4:4:2$ when preparing the dataset):
\begin{equation}
p^{\textit {true }}(\textit{\uppercase\expandafter{\romannumeral1}}>\textit{\uppercase\expandafter{\romannumeral2}})= \begin{cases}1 & \text { if } \mu_\textit{\uppercase\expandafter{\romannumeral1}}^{\textit{true}}-\mu_\textit{\uppercase\expandafter{\romannumeral2}}^{\textit{true}}> {\sqrt{\left(\sigma_\textit{\uppercase\expandafter{\romannumeral1}}\right)^2+\left(\sigma_\textit{\uppercase\expandafter{\romannumeral2}}\right)^2}}\\ 0 &\text { if } \mu_\textit{\uppercase\expandafter{\romannumeral2}}^{\textit{true}}-\mu_\textit{\uppercase\expandafter{\romannumeral1}}^{\textit{true}}> {\sqrt{\left(\sigma_\textit{\uppercase\expandafter{\romannumeral1}}\right)^2+\left(\sigma_\textit{\uppercase\expandafter{\romannumeral2}}\right)^2}}\\ 0.5 &\text { else (\textit{tie cases}) } .\end{cases}
\end{equation}
Thus, we define the predicted \textit{pairwise PMOD} as $P_{\textit {pred }}=[p^{\textit {pred}}(\textit{\uppercase\expandafter{\romannumeral1}}>\textit{\uppercase\expandafter{\romannumeral2}}),1-p^{\textit {pred }}(\textit{\uppercase\expandafter{\romannumeral1}}>\textit{\uppercase\expandafter{\romannumeral2}})]$ and the label \textit{pairwise PMOD} as $P_{\textit {true }}=[p^{\textit {true }}(\textit{\uppercase\expandafter{\romannumeral1}}>\textit{\uppercase\expandafter{\romannumeral2}}), 1-p^{\textit {true }}(\textit{\uppercase\expandafter{\romannumeral1}}>\textit{\uppercase\expandafter{\romannumeral2}})]$.
Since the prefix preceding the quality token can be easily fitted (with Eq.~(\ref{single})), the \textit{CE} loss is unnecessary during pairwise training. Therefore, we employ only the \textit{KL divergence} between the predicted and label \textit{pairwise PMOD} as the training loss: 
\begin{equation}
    \mathcal{L_\textit{PMOD-pair}} = \mathcal{D_{KL}}(P_{\textit {true }} \| P_{\textit {pred}}).
\end{equation}

\begin{table*}[t]\tiny
    \centering
    \renewcommand\arraystretch{1}
    \renewcommand\tabcolsep{2.5pt}
    \belowrulesep=0pt\aboverulesep=0pt

    \caption{Performance on the video quality scoring task. Datasets marked in \textit{italics} denote \textit{OOD} datasets. \textit{Mini-VQA} refers to \textit{Minimalistic-VQA}. \textit{DNN-Based} models and \textit{In-domain LMMs} are trained on human-labeled \textit{LSVQ (train)}. \textit{General LMMs} are used as the reference models of our \textit{VITAL-Series} with zero-shot inference. [Per column: highest in \textbf{\textcolor{red}{red}}, second in \textbf{\textcolor{blue}{blue}}, third in \textbf{boldface}.]}
    \vspace{-10pt}
    \resizebox{\linewidth}{!}{%
    \begin{tabular}{l|ccccccccccccccccc}
    \hline
    \multicolumn{1}{l|}{\textbf{Datasets}}
      & \multicolumn{2}{c}{\textbf{LSVQ(1080p)}}
      & \multicolumn{2}{c}{\textbf{LSVQ(test)}}
      & \multicolumn{2}{c}{\textbf{LIVE-VQC}}
      & \multicolumn{2}{c}{\textbf{KoNViD-1K}}
      & \multicolumn{2}{c}{\textbf{YT-UGC}}
      & \multicolumn{2}{c}{\textbf{\textit{YT-Gaming}}}
      & \multicolumn{2}{c}{\textbf{\textit{CGVDS}}}  & \multicolumn{2}{c}{\textbf{\textit{KVQ}}}&\multirow{3}{*}{\textbf{AVG.$\uparrow$}}\\
    \cline{1-17}
    \multicolumn{1}{l|}{\textbf{\# of videos}}
      & \multicolumn{2}{c}{\textbf{3,573}}
      & \multicolumn{2}{c}{\textbf{7,182}}
      & \multicolumn{2}{c}{\textbf{585}}
      & \multicolumn{2}{c}{\textbf{1,200}}
      & \multicolumn{2}{c}{\textbf{1,098}}
      & \multicolumn{2}{c}{\textbf{600}}
      & \multicolumn{2}{c}{\textbf{357}}  & \multicolumn{2}{c}{\textbf{2,926}}&\\
    \cline{1-17}
      \textbf{Models} & SRCC$\uparrow$ & PLCC$\uparrow$
      & SRCC$\uparrow$ & PLCC$\uparrow$
      & SRCC$\uparrow$ & PLCC$\uparrow$
      & SRCC$\uparrow$ & PLCC$\uparrow$
      & SRCC$\uparrow$ & PLCC$\uparrow$
      & SRCC$\uparrow$ & PLCC$\uparrow$ &SRCC$\uparrow$ & PLCC$\uparrow$&SRCC$\uparrow$ & PLCC$\uparrow$ &\\
    \cdashline{1-18}
      \multicolumn{1}{l}{\textit{DNN-Based}}\\ 
      \cdashline{1-18}
    \textsc{FAST-VQA (TPAMI 2023)}~\cite{wu2023neighbourhood}
      & 0.765 & 0.793
      & 0.880 & 0.871
      & \textbf{\textcolor{blue}{0.830}} & 0.822
      & 0.869 & 0.870
      & 0.725 & 0.742
      & 0.631 & 0.677
      & 0.725 & 0.747 & 0.518 &0.526 &0.749 \\
     \textsc{Mini-VQA-VII (TPAMI 2024)}~\cite{sun2024analysis}
      &0.740 &0.784
      &0.861 &0.859
       &0.757 &0.813
      & 0.843 &0.841
      & 0.775 &0.779
      & 0.596 &0.682
      & 0.681 &0.733
      & 0.604 &0.659 &0.750\\
    \textsc{DOVER (ICCV 2023)}~\cite{wu2023exploring}
      & \textbf{\textcolor{blue}{0.797}} & \textbf{0.821}
      & \textbf{\textcolor{blue}{0.893}} & \textbf{\textcolor{blue}{0.892}}
      & \textbf{\textcolor{red}{0.835}} & \textbf{\textcolor{red}{0.857}}
      & \textbf{\textcolor{blue}{0.885}} & 0.879
      & 0.801 & 0.814
      & 0.647 & 0.728
      &0.694 &0.747 &0.559 &0.593 &0.778 \\
    \textsc{KVQ (CVPR 2025)}~\cite{qu2025kvq}
      & \textbf{\textcolor{red}{0.814}} & \textbf{\textcolor{red}{0.846}}
      & \textbf{\textcolor{red}{0.896}} & \textbf{\textcolor{red}{0.897}}
      & \textbf{0.820} & \textbf{\textcolor{blue}{0.843}}
      & \textbf{\textcolor{red}{0.890}} & \textbf{\textcolor{red}{0.892}}
      & 0.829 & 0.818
      & 0.625 & 0.632
      &0.754 &0.792 &0.559 &0.576 &0.780 \\
      \cdashline{1-18}
      
      \multicolumn{1}{l}{\textit{In-domain LMMs}}\\ 
      \cdashline{1-18}
    \textsc{Q-Align (ICML 2024)} ~\cite{wu2024q1}
      & \textbf{\textcolor{blue}{0.797}} & \textbf{\textcolor{blue}{0.830}}
      & \textbf{0.883} & \textbf{0.882}
      & 0.773& 0.829
      & 0.865 & 0.877
      & 0.831 & \textbf{0.847}
      & 0.611 & 0.681
      & 0.756 &0.798 &0.555 &0.606 &0.776 \\
    \textsc{VQA-UGC-Scorer (MM 2025)} ~\cite{jia2024vqa}
      & 0.794 & \textbf{0.821}
      & 0.878 & 0.872
      & 0.785 & 0.830
      & \textbf{0.881} & \textbf{0.880}
      & 0.811 & 0.823
      & 0.613 & 0.698
      & 0.656 &0.741
      & 0.678 &0.689&0.778\\
    \textsc{Internvl3-SFT-8B-\textit{LSVQ}}~\cite{zhu2025internvl3}
     &0.749&0.777&0.849&0.839&0.743&0.790&0.846&0.843&0.798&0.800&0.632&0.693&0.801&0.808&0.560&0.592 &0.758 \\
    \cdashline{1-18}
    \multicolumn{1}{l}{\textit{General LMMs (reference)}}\\ 
      \cdashline{1-18}
      \textsc{Internvl3-Instruct-1B}
      &0.437&0.477&0.352&0.382&0.422&0.476&0.516&0.538&0.480&0.511&0.276&0.282&0.338&0.388&0.285&0.312 &0.405 \\
      \textsc{Internvl3-Instruct-2B}&0.468&0.492&0.463&0.468&0.418&0.434&0.575&0.563&0.510&0.495&0.301&0.291&0.114&0.149&0.287&0.297&0.395
      \\
      \textsc{Internvl3-Instruct-8B} &0.523&0.453&0.571&0.485&0.411&0.363&0.512&0.453&0.595&0.520&0.293&0.294&0.263&0.145&0.278&0.257&0.401
      \\
      
      \textsc{Internvl3-Instruct-14B} &0.499&0.562&0.602&0.582&0.374&0.413&0.580&0.587&0.596&0.601&0.241&0.268&0.293&0.361&0.365&0.335&0.454
      \\ 
    \cdashline{1-18}
      \multicolumn{1}{l}{\textit{VITAL-Series}}\\ 
      \cdashline{1-18}
    \textsc{VITAL-Base-8B}
      & 0.786&0.815&\textbf{0.883}&0.879&0.800&\textbf{\textcolor{blue}{0.843}}&0.878&\textbf{\textcolor{blue}{0.881}}&\textbf{\textcolor{red}{0.854}}&\textbf{\textcolor{red}{0.856}}&\textbf{\textcolor{red}{0.710}}&\textbf{\textcolor{red}{0.764}}&\textbf{\textcolor{red}{0.830}}&\textbf{\textcolor{red}{0.854}}&\textbf{0.721}&\textbf{\textcolor{red}{0.766}}&\textbf{\textcolor{red}{0.820}}\\
    \textsc{VITAL-Zero-1B}  &0.655&0.636&0.715&0.709&0.665&0.670&0.672&0.680&0.643&0.637&0.481&0.503&0.679&0.674&0.428&0.330&0.611
     \\ 
    \textsc{VITAL-Zero-2B} &0.660&0.686&0.765&0.750&0.688&0.713&0.783&0.747&0.676&0.655&0.582&0.615&0.759&0.693&0.562&0.595&0.683
       \\ 
       \textsc{VITAL-Zero-14B} &0.663&0.689&0.802&0.732&0.719&0.729&0.782&0.717&0.790&0.732&0.567&0.604&0.766&0.761&0.553&0.505&0.694 \\
    \textsc{VITAL--Warm-up-1B}
      & 0.786&0.819&0.869&0.866&0.771&0.812&0.873&0.875&\textbf{\textcolor{blue}{0.846}}&\textbf{\textcolor{blue}{0.848}}&\textbf{\textcolor{blue}{0.705}}&\textbf{\textcolor{blue}{0.754}}&\textbf{\textcolor{blue}{0.808}}&\textbf{\textcolor{blue}{0.824}}&\textbf{\textcolor{blue}{0.730}}&\textbf{0.743}&\textbf{\textcolor{blue}{0.808}}\\
    \textsc{VITAL-Warm-up-2B} 
    &0.787&0.817&0.869&0.866&0.768&0.816&0.874&0.878&\textbf{0.843}&0.844&\textbf{0.700}&\textbf{0.750}&\textbf{0.807}&\textbf{0.819}&0.654&0.669&\textbf{\textbf{0.798}}\\
    \textsc{VITAL-Warm-up-14B}
      & 0.787&0.806&0.870&0.865&0.773&0.819&0.875&0.877&\textbf{0.843}&0.841&0.688&0.722&0.804&0.815&\textbf{\textcolor{red}{0.740}}&\textbf{\textcolor{blue}{0.765}}&0.795 \\
    
    \cline{1-18}
    \end{tabular}%
    }
    \vspace{-1pt}
    \label{tab:VQA}
\end{table*}

\begin{table*}[tbp]
\vspace{-1em}
\centering
\small
\renewcommand\arraystretch{0.95}
\renewcommand\tabcolsep{8.5pt}
\caption{Performance on the image quality scoring task. Datasets marked in \textit{italics} denote \textit{OOD} datasets. \textit{DNN-Based} models and \textit{In-domain LMMs} are trained on human-labeled \textit{KonIQ (train)}.}
\vspace{-10pt}
\resizebox{\linewidth}{!}{%
\begin{tabular}{l|ccccccccccccccc}
\cline{1-16}
\textbf{Datasets}
  & \multicolumn{2}{c}{\textbf{KonIQ}}
  & \multicolumn{2}{c}{\textbf{SPAQ}}
  & \multicolumn{2}{c}{\textbf{LIVE-C}}
  & \multicolumn{2}{c}{\textbf{\textit{AGIQA}}}
  & \multicolumn{2}{c}{\textbf{\textit{KADID}}}& \multicolumn{2}{c}{\textbf{\textit{TID}}}&\multicolumn{2}{c}{\textbf{\textit{CSIQ}}}&\multirow{3}{*}{\textbf{AVG.$\uparrow$}}
 \\ 
\cline{1-15}
\textbf{\# of images}
  & \multicolumn{2}{c}{\textbf{2,010}}
  & \multicolumn{2}{c}{\textbf{2,224}}
  & \multicolumn{2}{c}{\textbf{1,169}}
  & \multicolumn{2}{c}{\textbf{2,982}}
  & \multicolumn{2}{c}{\textbf{2,000}}& \multicolumn{2}{c}{\textbf{3,000}}&\multicolumn{2}{c}{\textbf{866}}&
 \\ 
\cline{1-15}
  \textbf{Models}& SRCC$\uparrow$ & PLCC$\uparrow$ & SRCC$\uparrow$ & PLCC$\uparrow$ & SRCC$\uparrow$ & PLCC$\uparrow$ & SRCC$\uparrow$ & PLCC$\uparrow$ & SRCC$\uparrow$ & PLCC$\uparrow$ & SRCC$\uparrow$ & PLCC$\uparrow$ & SRCC$\uparrow$ & PLCC$\uparrow$ &  \\
\cdashline{1-16}
      \multicolumn{1}{l}{\textit{DNN-Based}}\\ 
\cdashline{1-16}
\textsc{NIMA (TIP 2018)} ~\cite{nima}
  & 0.859 & 0.896 
  & 0.856 & 0.838 
  & 0.771 & 0.814 
  & 0.654 & 0.715 
  & 0.535 & 0.532 
  & 0.381 & 0.426
  &0.604 &0.607&0.678\\

\textsc{DBCNN (TCSVT 2020)} ~\cite{dbcnn}
  & 0.875 & 0.884 
  & 0.806 & 0.812 
  & 0.755 & 0.773 
  & 0.641 & 0.730 
  & 0.484 & 0.497 
  & 0.385 &0.514 
  & 0.572 &0.573&0.664\\

\textsc{HyperIQA (CVPR 2020)} ~\cite{hyperiqa}
  & 0.906 & 0.917 
  & 0.788 & 0.791 
  & 0.749 & 0.772 
  & 0.640 & 0.702 
  & 0.468 & 0.506 
  & 0.632 &0.701
  & 0.719 &0.759&0.718\\

\textsc{MUSIQ (ICCV 2021)} ~\cite{musiq}
  & 0.929 & 0.924 
  & 0.863 & 0.868 
  & 0.830 & 0.789 
  & 0.630 & 0.722 
  & 0.556 & 0.575 
  & 0.629 &\textbf{\textcolor{blue}{0.704}} 
  & 0.721 &0.754&0.750\\

\textsc{TReS (WACV 2022)} ~\cite{golestaneh2022no}
  &0.916 &0.926
  &0.860 &0.861
  & 0.771 &0.805
  & 0.623 &0.716
  & 0.468 & 0.492
  & 0.393 &0.508
  & 0.572 &0.573&0.677
  \\

\textsc{CLIP-IQA+ (AAAI 2023)} ~\cite{wang2023exploring}
  & 0.895 & 0.909 
  & 0.864 & 0.866 
  & 0.805 & 0.832 
  & 0.685 & 0.736 
  & 0.654 & 0.653 
  & 0.575 &0.681 
  & 0.710 &0.770&0.760  \\

\textsc{LIQE (CVPR 2023)} ~\cite{zhang2023blind}
  & 0.928 & 0.912 
  & 0.833 & 0.846 
  & \textbf{\textcolor{blue}{0.870}} & 0.830 
  & 0.708 & 0.772 
  & 0.662 & 0.667 
  & 0.658 &\textbf{0.703}
  &\textbf{\textcolor{blue}{0.808}} &0.782&0.784\\

\textsc{TOPIQ (TIP 2024)} ~\cite{topiq}
  &0.930 & \textbf{\textcolor{blue}{0.944}} & 0.870 &0.874& 0.811 &0.826 &0.720&\textbf{0.796}&0.511 &0.546 &0.445 &0.562 &0.748 &0.767&0.739\\
  \cdashline{1-16}
      
      \multicolumn{1}{l}{\textit{In-domain LMMs}}\\ 
      \cdashline{1-16}
\textsc{Q-Align} ~\cite{wu2024q1}
  &\textbf{\textcolor{blue}{0.940}} & \textbf{0.941} 
  &\textbf{\textcolor{blue}{0.887}} &\textbf{\textcolor{blue}{0.886}}
  & \textbf{0.860} & 0.853
  & \textbf{\textcolor{blue}{0.735}} & 0.772
  &0.684 & \textbf{0.674}
  & 0.568& 0.671  &
  0.737& 0.785&0.785\\

\textsc{DeQA-Score (CVPR 2025)} ~\cite{you2025teaching}
  &\textbf{\textcolor{red}{0.941}} & \textbf{\textcolor{red}{0.953}}
  & \textbf{\textcolor{red}{0.896}}& \textbf{\textcolor{red}{0.895}}
  & \textbf{\textcolor{red}{0.879}} & \textbf{\textcolor{red}{0.892}} 
  & \textbf{0.729} & \textbf{\textcolor{blue}{0.809}} 
  & \textbf{{0.687}} & \textbf{\textcolor{blue}{0.694}} 
  & 0.607 &{0.673}& 0.744 &0.787&\textbf{\textcolor{blue}{0.799}} \\

\textsc{Internvl3-SFT-8B-\textit{KonIQ}}~\cite{zhu2025internvl3}
  &0.913&0.914&0.863&0.870&0.817&0.822&0.678&0.773&0.521&0.553&0.459&0.565&0.762&0.790&0.736\\
\cdashline{1-16}
\multicolumn{1}{l}{\textit{General LMMs (Reference)}}\\ 
      \cdashline{1-16} 
      \textsc{Internvl3-Instruct-1B}&0.630&0.672&0.750&0.757&0.625&0.648&0.566&0.602&0.410&0.404&0.417&0.434&0.452&0.511&0.563
    \\ 
      \textsc{Internvl3-Instruct-2B} &0.698&0.671&0.605&0.603&0.667&0.637&0.683&0.686&0.490&0.500&0.524&0.483&0.596&0.573&0.601
     \\
      \textsc{Internvl3-Instruct-8B} &0.766&0.769&0.623&0.698&0.705&0.654&0.686&0.791&0.579&0.573&0.536&0.491&0.650&0.596&0.658
     \\ 
      \textsc{Internvl3-Instruct-14B} &0.737&0.701&0.560&0.652&0.618&0.586&0.655&0.674&0.536&0.541&0.478&0.498&0.509&0.496&0.589
      \\
\cdashline{1-16}
      \multicolumn{1}{l}{\textit{VITAL-Series}}\\ 
      \cdashline{1-16}
    \textsc{VITAL-Base-8B}
&\textbf{0.931}&0.931&\textbf{0.884}&\textbf{\textcolor{blue}{0.886}}&0.851&\textbf{\textcolor{blue}{0.866}}&\textbf{\textcolor{red}{0.736}}&\textbf{\textcolor{red}{0.811}}&\textbf{\textcolor{red}{0.759}}&\textbf{\textcolor{red}{0.708}}&\textbf{\textcolor{red}{0.680}}&\textbf{\textcolor{red}{0.707}}&\textbf{\textcolor{red}{0.823}}&\textbf{\textcolor{red}{0.851}}&\textbf{\textcolor{red}{0.816}}\\
    \textsc{VITAL-Zero-1B}&0.819&0.744&0.765&0.762&0.649&0.610&0.647&0.665&0.614&0.636&0.663&0.650&0.650&0.653&0.681\\
    \textsc{VITAL-Zero-2B} &0.897&0.855&0.862&0.856&0.810&0.783&0.725&0.738&0.625&0.630&0.662&0.648&0.764&0.781&0.760 \\
    \textsc{VITAL-Zero-14B} &0.876&0.878&0.868&0.841&0.800&0.765&0.709&0.721&\textbf{\textcolor{blue}{0.706}}&0.665&0.610&0.652&0.774&0.765&0.759 \\
    \textsc{VITAL--Warm-up-1B}
      & 0.918&0.917&0.871&0.877&0.851&\textbf{0.859}&0.716&0.783&0.674&0.659&\textbf{0.674}&0.667&\textbf{{0.787}}&\textbf{0.793}&0.789 \\
    \textsc{VITAL-Warm-up-2B} &0.919&0.916&0.873&0.878&0.852&\textbf{0.859}&0.719&0.781&0.674&0.646&0.652&0.663&0.782&0.792&0.786\\
    \textsc{VITAL-Warm-up-14B}
      & 0.918&0.917&0.870&0.875&0.853&0.857&0.720&0.784&0.670&0.645&\textbf{\textcolor{blue}{0.676}}&0.691&0.786&\textbf{\textcolor{blue}{0.796}}&\textbf{0.791}\\
    
\cline{1-16}
\end{tabular}}
\vspace{-5pt}
\label{tab:IQA}
\end{table*}
\paragraph{Text Generation Training.}
The text generation training consists of two subtasks: distortion identification and visual quality description. 
The training corpus comprises statements exhibiting diverse lengths and linguistic complexity. For relatively short and semantically simple descriptions, such as those specifying particular distortion types and severity, the model tends to converge more readily, reflected by a rapid increase in output token probabilities. 
In contrast, longer and semantically richer statements pose greater representational challenges, causing the model to exhibit slower or less pronounced probability gains across target tokens.
Under such conditions, the model tends to overfit toward shorter and simpler output forms. To handle this issue, we adopt the \textit{focal loss}~\cite{lin2017focal} that dynamically adjusts the learning rate for each output token based on its instantaneous target token output probability. This effectively mitigates the imbalance between simple and complex VL pairs by emphasizing tokens that are harder to predict and down-weighting those that are already well fitted:
\begin{equation}
\mathcal{L_\textit{Interp}} = -\frac{1}{L}\left(\sum_{\ell=0}^{L-1}\alpha (1 - p(\mathbf{z_{\ell}} | \mathbf{Z_{\ell}}))^\beta \log p(\mathbf{z_{\ell}} | \mathbf{Z_{\ell}})\right),
\end{equation}
where $\alpha=1$ and $\beta=2$.

\paragraph{Pre-training Details.}
All the training data is randomly mixed. We set the \textsc{epoch-num} to $1$ and \textsc{batch-size-per-device} to $2$. The training requires approximately $1920$ GPU hours using 8-GPU H200. After training, we refer to the full LMM—comprising the \textit{VITAL Vision Encoder} and the other frozen part (the \textbf{homogeneous decoder})—as \textit{VITAL-Base-8B}. Moreover, to mitigate potential biases in model outputs induced by identical textual guidance, we employ the \textbf{prompt disentanglement} strategy. Specifically, we rely exclusively on the vision token sequence as the input tokens, excluding any textual guides. This setting enables the LMM to directly evoke visual quality
understanding through the vision tokens themselves, avoiding overfitting to frequently encountered textual prefixes.


\subsection{Efficient Model Zoo Extension}

General LMMs typically undergo extensive pre-training, during which they acquire and share rich VL priors.
 This representational foundation provides the practical basis for \textbf{efficient model zoo extension} (in Fig.~\ref{fig:workflow}). Here, we utilize the \textit{VITAL Visual Encoder} as the base socket and integrate it with diverse decoders to construct a model zoo encompassing variations in functionality and structure.

For the \textit{VITAL-Base-8B} with the \textbf{homogeneous decoder}, we further enhance it via post-training to improve its comprehensiveness in visual quality interpreting. To this end, we aggregate a collection of publicly available MIDBs (including \textit{Q-Pathway-200K}, \textit{AesMMIT-400K}, \textit{VQA\textsuperscript{2}-Stage3-DB-115K}, and \textit{OmniVQA-Chat-400K} with a total of $1120K$ instruction VL pairs) and apply full-parameter SFT using the \textit{focal loss}, yielding the \textit{VITAL-Assistant-8B}.

For \textbf{heterogeneous decoders} (here we choose the LLM in \textit{InternVL-1B}, \textit{InternVL-2B}, and \textit{InternVL-14B} with their corresponding projectors) that are not used during pre-training, we propose \textbf{two transfer strategies}. (1) The pre-trained \textit{VITAL Visual Encoder} is directly combined with the target decoder, producing the \textit{VITAL-Zero} series. (2) After transferring the decoder, we sample $4000$ instances from the pre-training data (preserving the original task distribution) and perform efficient decoder-only \textbf{warm-up training}, resulting in the \textit{VITAL-Warm-up} series. 

Extensive experiments justify that all models deliver satisfactory performance on their respective VQualA tasks, underscoring the strong transferability.

%% file: sec/4_experiments.tex
\vspace{-5pt}
\section{Experiments}

\begin{table*}[t]\tiny
    \centering
    \renewcommand\arraystretch{1.05}
    \renewcommand\tabcolsep{9pt}
    \belowrulesep=0pt\aboverulesep=0pt

    \caption{Evaluation results on the \textit{Qbench-video-test-single}. \textit{Multi.} refers to the question type with four options and a single answer.
}
 \vspace{-10pt}
    \resizebox{\linewidth}{!}{\begin{tabular}{l|ccc|cccc|c}
  
       \hline
     \textbf{Sub-categories} & \multicolumn{3}{c|}{\textbf{Question Types}} & \multicolumn{4}{c|}{\textbf{Quality Concerns}} &\multirow{2}{*}{\textit{Overall$\uparrow$}}
        \\ \cline{1-8}
        \multirow{1}{*}{\textbf{Models}} & \textit{Binary} $\uparrow$&\textit{Multi.} $\uparrow$& \textit{Open-ended} $\uparrow$& \textit{Technical} $\uparrow$&\textit{Aesthetic} $\uparrow$& \textit{Temporal}  $\uparrow$&\textit{AIGC} $\uparrow$& \\
    \hdashline
      \multicolumn{1}{l}{\textit{Open-sourced General LMMs}}\\ 
      \hdashline
      \textsc{InternVL3-8B (base model)}  ~\cite{zhu2025internvl3} &50.84\% & 44.25\% & 33.54\% & 41.82\% & 52.37\% & 45.07\% & 37.58\% & 42.67\%\\ 
      \textsc{InternVL3.5-8B} ~\cite{wang2025internvl3} &56.23\% & 49.83\% & 38.61\% & 47.33\% & 58.77\% & 51.36\% & 38.51\% & 48.00\% \\ 
       \textsc{InternVL3.5-14B}&55.89\% & 48.43\% & 39.56\% & 45.22\% & 55.92\% & 51.36\% & 42.55\% & 47.78\%\\ 
        \textsc{InternVL3.5-38B}&56.23\% & 48.08\% & 41.14\% & 46.03\% & 58.06\% & 50.51\% & 43.17\% & 48.33\%\\ 
        \textsc{Qwen3vl-4B}~\cite{xu2025qwen3}&54.21\% & 46.34\% & 45.09\% & 48.22\% & 59.95\% & 50.85\% & 36.65\% & 48.50\%\\ 
         \textsc{Qwen3vl-8B}&56.57\% & 46.34\% & 42.56\% & 47.24\% & 59.95\% & 49.49\% & 41.30\% & 48.39\%\\ 
           \textsc{Qwen3vl-32B}&64.31\% & 54.70\% & 43.51\% & 52.43\% & \textbf{64.93\%} & 56.97\% & 45.03\% & 53.94\%\\ 
            \hdashline
        \multicolumn{1}{l}{\textit{Domain Specific LMMs}}\\ 
        \hdashline
        \textsc{VQA\textsuperscript{2}-Assistant} ~\cite{jia2024vqa}&67.12\% & 59.93\% & 39.56\% & 55.19\% & 56.87\% & \textbf{57.99\%} & 43.79\% & 55.56\%\\ 
       \textsc{OmniVQA-Chatter} ~\cite{jia2025scaling}&\textbf{68.69\%} & \textbf{\textcolor{blue}{68.29\%}} & 44.15\% & \textbf{58.67\%} & 62.56\% & \textbf{\textcolor{blue}{60.54\%}} & \textbf{53.73\%} & \textbf{59.94\%}\\ 
       \hdashline
      \multicolumn{1}{l}{\textit{Proprietary LMMs}}\\ 
      \hdashline
      \textsc{GPT-4o (24-11-20)} ~\cite{achiam2023gpt} &59.26\% & 51.57\% & \textbf{47.63\%} & 51.38\% & 61.37\% & 51.87\% & 49.38\% & 52.72\%\\ 
       \textsc{GPT-5 (25-08-07)} &56.57\% & 55.05\% & \textbf{\textcolor{blue}{48.73\%}} & 52.03\% & \textbf{\textcolor{blue}{66.11\%}} & 55.78\% & 51.55\% & 53.33\%\\ 
      \textsc{Gemini-2.5-Pro} ~\cite{mallick2025gemini} &\textbf{\textcolor{blue}{69.70\%}} & \textbf{63.07\%} & \textbf{\textcolor{red}{54.75\%}} & \textbf{\textcolor{blue}{61.43\%}} & \textbf{\textcolor{red}{68.48\%}} & 56.97\% & \textbf{\textcolor{red}{66.46\%}} & \textbf{\textcolor{blue}{62.33\%}} \\ 
      \textsc{Qwen-VL-Max (25-08-13)}~\cite{bai2023qwen} &56.57\% & 55.05\% & 43.99\% & 50.49\% & 64.45\% & 50.34\% & 49.69\% & 51.67\% \\ 
      \hdashline
      \multicolumn{1}{l}{\textit{VITAL-Series}}\\ 
      \hdashline
      \textsc{VITAL-Base-8B}&61.28\% & 58.19\% & 35.76\% & 52.43\% & 53.08\% & 50.34\% & 41.93\% & 51.33\%\\ 
      \textsc{VITAL-Assistant-8B}&\textbf{\textcolor{red}{72.05\%}} & \textbf{\textcolor{red}{72.13\%}} & {46.52\%} & \textbf{\textcolor{red}{61.75\%}} & {64.45\%} & \textbf{\textcolor{red}{61.56\%}} & \textbf{\textcolor{blue}{60.25\%}} & \textbf{\textcolor{red}{63.11\%}}\\ 
     
      \cline{1-9}
      \end{tabular}}
      \vspace{-12pt}

    \label{tab:interp}
\end{table*}

To evaluate the performance and transferability of the \textit{VITAL-Series}, we conduct comparison experiments across multiple key tasks. Moreover, our evaluation also highlights the discussions and ablation studies of key attributes.
\subsection{Experiments Settings}

For VQA-scoring, we evaluate on $8$ open-sourced datasets: \textit{LSVQ (test)}~\cite{ying2021patch}, \textit{LSVQ (1080P)}, \textit{KoNViD-1K}~\cite{hosu2017konstanz}, \textit{LIVE-VQC}~\cite{ghadiyaram2017capture}, \textit{YouTube-UGC}~\cite{wang2019youtube}, \textit{LIVE-YouTube-Gaming}~\cite{yu2022subjective}, \textit{CGVDS}~\cite{saha2023study}, and \textit{KVQ}~\cite{lu2024kvq}. These datasets collectively cover diverse scenarios.  We compare against multiple SOTA methods based on deep neural network (DNN) or LMM (in Tab.~\ref{tab:VQA}). All compared models use publicly released weights trained on the \textit{LSVQ (train)} ($28056$ samples).  For IQA-scoring, we utilize $7$ representative datasets: \textit{KonIQ-10K (test)}~\cite{hosu2020koniq}, \textit{SPAQ}~\cite{fang2020perceptual}, \textit{KADID-10K}~\cite{lin2019kadid}, \textit{AGIQA-3K}~\cite{li2023agiqa}, \textit{TID-2013 (TID)}~\cite{ponomarenko2015image}, \textit{CSIQ}~\cite{larson2010most}, and \textit{LIVE-Challenge (LIVE-C)}~\cite{livechallenge}. Similarly, we include various high-performing methods for comparison (in Tab.~\ref{tab:IQA}). All compared models are trained on \textit{KonIQ-10K (train)} ($7046$ samples) by default. Furthermore, since our pre-training strategy follows an \textit{opinion-unaware (OU)} paradigm requiring no human-annotated labels for training, we conduct comparison experiments on the IQA-scoring task against several recent \textit{OU} approaches. All baselines are implemented according to their official open-sourced configurations to ensure experimental fairness. The results are depicted in Tab.~\ref{tab:OU}. We use two widely used metrics for evaluation: the \textit{Pearson Linear Correlation Coefficient (PLCC)} and the \textit{Spearman Rank Correlation Coefficient (SRCC)}.


For the text generation task, we evaluate on the \textit{QBench-video-test-single}~\cite{zhang2024q}, which consists of $900$ questions targeting various video and distortion types with multiple quality concerns. Alongside our models, we include both recently released open-source and proprietary general and in-domain LMMs as baselines to enhance the credibility of the experimental results (demonstrated in Tab.~\ref{tab:interp}).

During \textbf{inference}, we generally follow the settings of \textit{VQA\textsuperscript{2}} with a little modification (detailed in \textit{Supp. Sec.~\ref{inference}}). 


\subsection{Main Results Analysis}

In the scoring task, the \textit{VITAL-Base-8B} achieves outstanding performance across all IQA and VQA datasets. It surpasses both its base model and the reference SFT model trained on human-annotated datasets (video: \textit{LSVQ (train)}; image: \textit{KonIQ (train)}). Furthermore, on most of the \textit{out-of-distribution} (\textit{OOD}) datasets (video: \textit{LIVE-YouTube-Gaming}, \textit{CGVDS}, and \textit{KVQ}; image: \textit{KADID-10k}, \textit{AGIQA-3K}, \textit{TID-2013}, and \textit{CSIQ}), \textit{VITAL-Base-8B} demonstrates clear superiority over the strongest baselines (VQA: \textit{KVQ}; IQA: \textit{DeQA-Score}). 
For the \textit{VITAL-Zero} series, despite the absence of fine-tuning, the models still achieve substantial performance gains compared to their base models. This underscores the crucial role of generative pre-training in achieving robust zero-shot performance. 
Moreover, the results of the \textit{VITAL-Warm-up} series demonstrate strong transferability. The \textit{1B}, \textit{2B}, and \textit{14B} models achieve consistently competitive scoring performance. Notably, their \textit{OOD} performance (in most cases) shows only marginal degradation compared with \textit{VITAL-Base-8B}.
In comparison with \textit{OU} methods, both the \textit{VITAL-Base} and \textit{VITAL-Warm-up} models achieve markedly higher performance. 
The above observations indicate that generative pre-training, in the absence of human annotations, can still significantly improve the VQualA LMM’s generalization and transferability to various types of visual content.

In the text generation task, the \textit{VITAL-Base-8B} largely preserves its instruction-following capability (through vision-encoder-centered training without tuning the LLM) and outperforms the base model.
 After post-training, the \textit{VITAL-Assistant-8B} achieves superior performance compared with both the latest general and domain-specific LMMs. We also provide a series of case studies (in \textit{Supp. Sec.~\ref{Case Studies}}) focused on real-world applications to further demonstrate the functionality of the \textit{VITAL-Assistant}.

\vspace{-2pt}

\subsection{Discussions}


\paragraph{Linear Probe Performance.}
Beyond compatibility with LLM decoders, verifying whether the \textit{VITAL Vision Encoder} can adapt to decoders with simpler architectures is also crucial to assessing its transferability. To this end, we employ the \textit{linear probe} setup. Specifically, we extract image feature tokens from different layers of \textit{InternViT} and motion tokens from \textit{SlowFast}, apply token-wise average pooling to each, concatenate the averaged feature tokens, and input them into the linear probe.
 We fine-tune the model on \textit{LSVQ (train)}. Consistent with the pre-training task, the linear probe outputs probabilities corresponding to five quality levels, and we use the \textit{CE loss} as the training objective. Detailed implementations are in \textit{Supp. Sec. \ref{Supplementary Experiments}}. We evaluate two training configurations: \textit{linear probe tuning} and \textit{full-param tuning}, and compare them against \textit{Simple-VQA}~\cite{sun2022deep} (with a similar architecture but without VQualA-specific pre-training) on the \textit{KoNViD-1k}, \textit{LIVE-VQC}, and \textit{LIVE-YT-Gaming}. Results are shown in Tab.~\ref{tab:linear}. We observe that even with lightweight linear probe tuning, the model surpasses the performance of the reference \textit{Simple-VQA}, while enabling full-param tuning further enhances its performance. This demonstrates the strong adaptability of the vision encoder within lightweight DNN decoders.
\begin{table}[t]\tiny
    \centering
    \renewcommand\arraystretch{0.9}
    \renewcommand\tabcolsep{6pt}
    \belowrulesep=0pt\aboverulesep=0pt
    \caption{Comparison with \textit{OU} IQA methods (SRCC~/~PLCC). [Per column: highest in \textbf{\textcolor{red}{red}}]}
    \vspace{-10pt}
    \resizebox{1\linewidth}{!}
    {\begin{tabular}{c|cccc}
    \hline
     \multicolumn{1}{c|}{\textbf{Models}}&{\textbf{KonIQ}} & \textbf{KADID}&  \textbf{SPAQ}&  \textbf{CSIQ} \\ 
     \hdashline
     \textit{OU Models}\\
     \hdashline
     \textsc{CLIP-IQA}~\cite{wang2023exploring}&0.642~/~0.709 & 0.501~/~0.520&0.733~/~ 0.734&0.681~/~0.716 \\
\textsc{ARNIQA}~\cite{agnolucci2024arniqa}&0.795~/~0.832& 0.725~/~0.717 & 0.788~/~0.797 &0.629~/~0.594 \\
      \textsc{QualiCLIP}~\cite{agnolucci2024quality}&0.715~/~0.765 & 0.665~/~0.665 & 0.841~/~0.851&0.781~/~0.815\\
      \hdashline
     \textit{VITAL-Series}\\
     \hdashline
       \textsc{Base-8B}&\textbf{\textcolor{red}{0.931}}~/~\textbf{\textcolor{red}{0.931}} & \textbf{\textcolor{red}{0.759}}~/~\textbf{\textcolor{red}{0.708}} &\textbf{\textcolor{red}{0.884}}~/~\textbf{\textcolor{red}{0.886}} &\textbf{\textcolor{red}{0.823}}~/~\textbf{\textcolor{red}{0.851}}\\
       \textsc{Warm-up-1B}&0.918~/~0.917 &0.674~/~0.659 & 0.871~/~0.877 &0.787~/~0.793 \\
       \textsc{Warm-up-2B}&0.919~/~0.916 & 0.674~/~0.646 &0.873~/~0.878 & 0.782~/~0.792 \\
       \textsc{Warm-up-14B}&0.918~/~0.917 &0.670~/~0.645 & 0.870~/~0.875 &0.786~/~0.796 \\
    \hline
    \end{tabular}}
     \vspace{-12pt}
    \label{tab:OU}
\end{table}

\begin{table}[t]\tiny
    \centering
    \renewcommand\arraystretch{1.10}
    \renewcommand\tabcolsep{5pt}
    \belowrulesep=0pt\aboverulesep=0pt
    \caption{Linear probe performance compared with the refering \textit{Simple-VQA}.}
    \vspace{-10pt}
    \resizebox{1\linewidth}{!}
    {\begin{tabular}{c|c|ccc}
    \hline
     \multicolumn{1}{c|}{\textbf{Version}}&\multicolumn{1}{c|}{\textbf{Tunable \#}}&{\textbf{LIVE-VQC}} &  \textbf{KoNViD-1k}&  \textbf{YT-Gaming} \\ \hline 
      \textsc{Linear-Probe}&$1.61M$ & 0.760~/~0.786 & 0.853~/~0.832 & 0.631~/~0.716  \\
      \textsc{Full}&$339.26M$ & \textbf{\textcolor{red}{0.780}}~/~\textbf{\textcolor{red}{0.808}} & \textbf{\textcolor{red}{0.870}}~/~\textbf{\textcolor{red}{0.873}} &\textbf{\textcolor{red}{0.692}}~/~\textbf{\textcolor{red}{0.735}}\\
    \hline
    \textsc{Simple-VQA}~\cite{sun2022deep}&$86.91M$& 0.749~/~0.789 & 0.826~/~0.820 & 0.656~/~0.717\\
    \hline 
    \end{tabular}}
     \vspace{-12pt}
    \label{tab:linear}
\end{table}

\begin{table}[t]\small
    \centering
    \renewcommand\arraystretch{1.10}
    \renewcommand\tabcolsep{2pt}
    \caption{Ablation study of key training attributes.}
    \vspace{-9pt}
    \resizebox{\linewidth}{!}{\begin{tabular}{c|ccc|ccc}
   
    \hline
    \multicolumn{1}{c|}{\textbf{\textsc{Attributes}}} &{\textbf{LIVE-VQC}} &  \textbf{KoNViD-1k}&  \textbf{YT-Gaming} &  \textbf{SPAQ} &  \textbf{KADID} &   \textbf{CSIQ} \\ 
    \hline
    \textsc{Base-8B} &\textbf{\textcolor{red}{0.800}}~/~\textbf{\textcolor{red}{0.843}} &\textbf{\textcolor{red}{0.878}}~/~\textbf{\textcolor{red}{0.881}} &0.710~/~\textbf{\textcolor{red}{0.764}} &\textbf{\textcolor{red}{0.884}}~/~\textbf{\textcolor{red}{0.886}} &\textbf{\textcolor{red}{0.759}}~/~0.708 &\textbf{\textcolor{red}{0.823}}~/~0.851\\
       \textsc{w/o PMOD} & 0.725~/~0.773 &0.835~/~0.840 &0.598~/~0.657 &0.865~/~0.872 &0.602~/~0.668 &0.743~/~0.785 \\
       \textsc{w/o Pair} & 0.785~/~0.817 &0.856~/~0.867 &0.703~/~0.760 &0.868~/~0.870 &0.725~/~0.687 &0.810~/~0.832\\
       \textsc{w/o Text} &0.790~/~0.837&0.868~/~0.873&\textbf{\textcolor{red}{0.712}}~/~0.760&0.876~/~0.880&0.743~/~\textbf{\textcolor{red}{0.712}}&0.818~/~\textbf{\textcolor{red}{0.854}}\\
    \hline
    \end{tabular}}
    \label{tab:ablation}
\end{table}

\begin{figure}[t]
    \centering
    \includegraphics[width=0.98\linewidth]{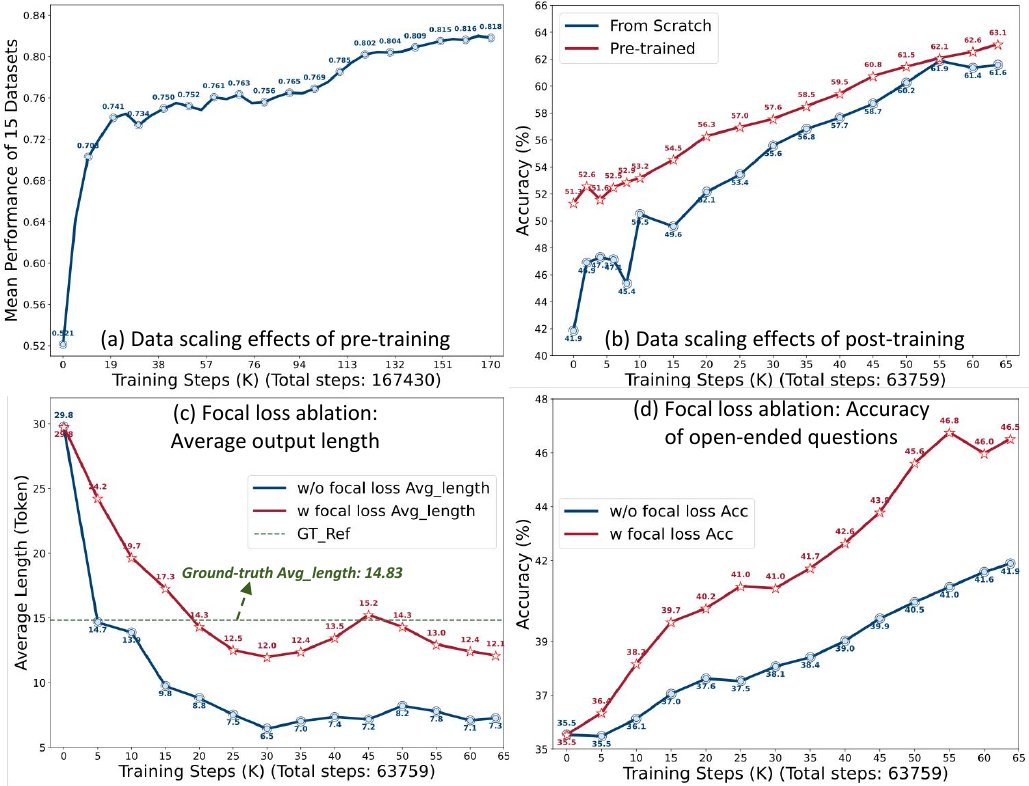}
    \vspace{-7pt}
    \caption{Plots of data scaling effects ((a) and (b)) and the focal loss ablation study ((c) and (d)).}

    \label{fig:plots}
\vspace{-6pt}   
\end{figure}

\paragraph{Data Scaling Effects.}
Recording the data scaling effects is essential for optimizing training efficiency. We measure how the average scoring performance of the \textit{VITAL-Base-8B} across the $15$ datasets varies as the pre-training data volume is proportionally scaled up. We also analyze the post-training data scaling effect for \textit{VITAL-Assistant}, referencing direct SFT on \textit{InternVL3-8B} (\textit{from-scratch}).
 The curves are displayed in Fig.~\ref{fig:plots} (a) and (b). For pre-training, the overall performance exhibits a positive correlation with the data volume, with performance gains gradually becoming marginal in the later training period. In post-training, both SFT strategies show an initial performance drop, while the pre-training-based approach remains much more stable, illustrating the domain generalization facilitated by pre-training.
Moreover, this strategy exhibits superior data efficiency compared to the \textit{from-scratch} approach. This further highlights the crucial role of text-generation pretraining in the post-training phase.

\paragraph{Key Training Attributes Ablation.}
We also conduct ablation studies to validate the contributions of various training tasks and tricks. Specifically, we report the results under different configurations: whether the \textit{PMOD} is used (otherwise, the model is trained only with a single visual input using the quality level corresponding to the mean machine opinion with \textit{CE loss}), whether \textit{pair-wise training} is included, and whether the \textit{text generation} tasks are involved. For all ablations, the training settings are kept identical to ensure fair comparison (results shown in Tab.~\ref{tab:ablation}). Results demonstrate that the \textit{PMOD} prediction serves as the primary contributor to the improvement in scoring performance. In addition, the incorporation of pairwise training further calibrates the scoring, while the inclusion of the text generation tasks also provides an additional enhancement in most cases.
To investigate the effect of \textit{focal loss}, we extract the $316$ open-ended questions in \textit{QBench-video-test-single}. Fig.~\ref{fig:plots} (c) and (d) present the comparison between the post-training with \textit{focal loss} and \textit{CE loss} in terms of the average output length and overall accuracy. Evidently, the model trained with focal loss consistently preserves longer output lengths and higher accuracy.

%% file: sec/5_conclusion.tex
\vspace{-5pt}
\section{Conclusion}
We construct a large-scale machine-annotated dataset comprising $4.58M$ VL pairs. Building upon this, we devise the \textbf{VITAL-Series}, a suite of \textbf{versatile}, \textbf{powerful}, and \textbf{transferable} VQualA LMMs, through the \textbf{vision-encoder-centered generative pre-training}. For quality scoring, our models significantly outperform SOTA baselines on \textit{OOD} scenarios. In text generation tasks, our post-trained model significantly outperforms various general and in-domain LMM competitors.
 Furthermore, the models exhibit strong structural transferability—demonstrating robust \textbf{zero-shot} and efficient \textbf{few-shot warm-up} performance when paired with diverse decoders. Overall, the VITAL-Series provides a new direction for developing \textbf{deployment-friendly} foundation LMMs tailored to VQualA.

%% file: sec/X_suppl.tex
\clearpage
\setcounter{page}{1}
\maketitlesupplementary

\section{Prompts Summary}
\label{prompt}
\subsection{Quality Scoring Task Prompts}

\begin{itemize}
    \item \textbf{Training:}  
    \textit{From Human:} [image]/[video]  
    \textit{From GPT:} The quality of the image/video is [level].
    
    \item \textbf{Inference:}  
    \textit{From Human:} [image]/[video] The quality of the image/video is  
    \textit{From GPT:} [Predicted Quality Level]
\end{itemize}
The difference between the training and inference prompts is designed to better locate the \textbf{quality token} and ensure the accuracy of the scoring prediction.

\subsection{Text Generation Task Prompts}

\begin{itemize}
    \item \textbf{Distortion Recognition Subtask}
    \begin{itemize}
        \item \textbf{Training:}  
        \textit{From Human:} [image]/[video]  
        \textit{From GPT:} [Distortion Severity]/[Distortion Category]
    \end{itemize}

    \item \textbf{Quality Interpreting Subtask}
    \begin{itemize}
        \item \textbf{Training:}  
        \textit{From Human:} [image]/[video]  
        \textit{From GPT:} [Corresponding Statement]
    
        \item \textbf{Annotator Instruction Prompts} 
        Describe the visual quality of the video/image in detail. /Elaborate on the quality of the video/image in detail.
    
        \item \textbf{Description Paragraph Processing Prompts:}  
        \begin{itemize}
            \item Please reformulate the paragraph into several concise sentence-level statements, split by periods. The requirements are as follows:
            \begin{itemize}
                \item First, discard all statements about vague descriptions of visual quality (not on explicit quality attributes).  
                \item Simultaneously, disregard any statements that provide a conclusion assessment of the video's visual quality (e.g., "thus, the visual quality of this video is high").
            \end{itemize}
            \item Directly output the revised description without any prefix or suffix.
            \item You should simulate as if you have derived the summary from the video itself, so do not reveal any trace of the provided description paragraph.
        \end{itemize}

        \item \textbf{Rejection Sampling Prompts}
        \begin{itemize}
            \item (Each round provides identical instructions to every judge. In practice, videos and images are processed sequentially, but for convenience, all are written in a single prompt here.)
            Please carefully observe the given video/image and assess whether you agree with the provided evaluation statement. Rate your assessment according to the following criteria:  
                \begin{itemize}
                \item 2 points: The evaluation statement is largely consistent with the given video/image, with only minor inaccuracies or non-standard descriptions. 
                \item 1 point: The evaluation statement shows some deviation from the key frame sequence, primarily due to inconsistencies in temporal quality or degree of distortions or noticeable inaccurate descriptions of quality factors.  
                \item 0 points: The evaluation statement is largely or completely inconsistent with the observed key frame sequence; the described elements do not appear in the given video/image or exhibit significant discrepancies. Please provide your reason if you rate 0 marks. 
                \item If you give a score of 1 point, please give your correction of the original statement, aiming to preserve a similar sentence structure and style, while making it more accurate and academic.
                \end{itemize}
        \end{itemize}
\end{itemize}
\begin{figure}[t]
    \centering
    \includegraphics[width=\linewidth]{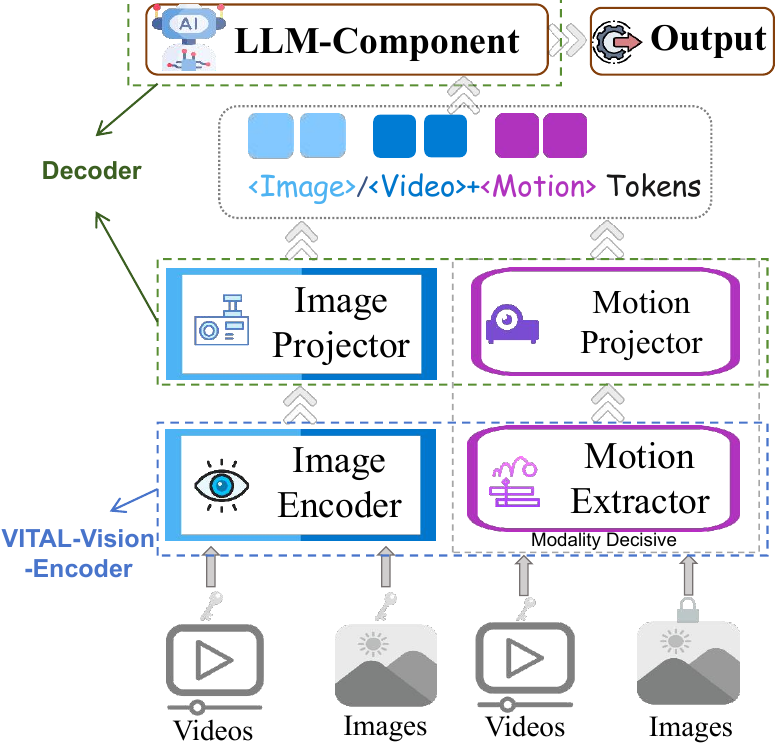}
    \caption{The model structure of the \textit{VITAL-Series}.}
    \label{fig:model}   
\end{figure}
    \item \textbf{Self-Judge Instruction Prompts}
    \begin{itemize}
        \item (Each round of instructions has a different structure, but the meaning remains consistent. The question is designed in a multiple-choice format to ensure that the output is verifiable and standardized.)
        \begin{itemize}
            \item Given one video/image, describe whether the visual quality assessment statement [statement] is correct.
            \begin{itemize}
                \item A. Yes  
                \item B. No
            \end{itemize}
        \end{itemize}
    \end{itemize}
\end{itemize}  

\section{Model Structure/Hyper-parameters}
The detailed model structure and hyperparameters are shown in Tab.~\ref{tab:hyperparam}. The model structure (with encoder/decoder split) and the ``prompt disentangled" input strategy are depicted in Fig.~\ref{fig:model}.

\label{structure}
\begin{table*}[h]
 \renewcommand\arraystretch{1.12}
\renewcommand\tabcolsep{6pt}
\belowrulesep=0pt\aboverulesep=0pt
\centering
\caption{
Details of the model structure and hyperparameters for the model training.  Entries without re-definition  indicate that the hyperparameter remains consistent.
}
\vspace{-10pt}
\resizebox{\linewidth}{!}{
\begin{tabular}{l| c| c}
\toprule
\textbf{Model Structure/Training Hyper-Parameters} &  \textbf{Name/Value} &  \textbf{More Information}  \\
\hline
\multicolumn{1}{l}{\textbf{VITAL-Base-8B / VITAL-Assistant-8B}} \\
\midrule
Vision encoder: Image encoder \textit{init.} & \textit{InternViT-300M-448px} &\textit{Parameter size=}$304.01$M\\
Vision encoder: Motion Extractor \textit{init.}& \textit{SlowFast-R50} &\textit{Parameter size}=$33.64$M,\textit{Use the fast-path feature}  \\
Decoder: Image projector \textit{init.} & \textit{2-layers MLP+GeLU}&\textit{Parameter size}=$27.54$M \textit{(Layernorm+Linear(1024,3584)+GELU+Linear(3584,3584))} \\
Decoder: Motion Projector \textit{init.}& \textit{2-layers MLP+GeLU}&\textit{Parameter size}=$13.77$M, \textit{(Layernorm+Linear(256,3584)+GELU+Linear(3584,3584))}  \\
Decoder: LLM \textit{init.} & \textit{Qwen-2.5-7B} &\textit{parameter size}=$7612.82$M,Decoder-only model\\
Keyframes Sampling Interval (For video) &1 second&/\\   
Video Keyframes / Image Patch Resolution         & $448 \times 448$&/\\
Token Feature Dimension (hidden size)         & $3584$&/\\

Frames (for motion extraction) Resolution         & $448\times 448$&/\\
Batch Size &$16$ ($8$ for pair-wise)&\textit{Per device train batch size=$2$ (for pair-wise training, this is set to $1$)} \\
LR Max & 2e-5 / 1e-5 &2e-5 in pre-traning, 1e-5 in post-training. \\
Gradient Accumulation Steps   & $2$&/ \\
Numerical Precision      & $\mathtt{bfloat16}$&/ \\
Epoch & $1$ & / \\
Eval Steps & None (no eval) & / \\
Optimizer & AdamW&/ \\
Activation Checkpointing &\checkmark&/ \\
Deepspeed Stage & $2$ &/ \\
\hline
\multicolumn{1}{l}{\textbf{VITAL-Warm-up-1B / VITAL-Zero-1B }} \\
\hline

Decoder: LLM \textit{init.}        & \textit{Qwen2.5-0.5B}&/\\
Token Feature Dimension (hidden size)        & $896$&/\\
\hline
\multicolumn{1}{l}{\textbf{VITAL-Warm-up-2B / VITAL-Zero-2B }} \\
\hline
Decoder: LLM \textit{init.}        & \textit{Qwen2.5-1.5B}&/\\
Token Feature Dimension  (hidden size)        & $1586$&/\\
\hline
\multicolumn{1}{l}{\textbf{VITAL-Warm-up-14B / VITAL-Zero-14B }} \\
\hline
Decoder: LLM \textit{init.}        & \textit{Qwen2.5-14B}&/\\
Token Feature Dimension (hidden size)        & $5120$&/\\
\bottomrule
\end{tabular}
}

\label{tab:hyperparam}

\end{table*}
\section{Additional Analysis for Main Methodology}
\subsection{Inference Details}
\label{inference}
\paragraph{Quality scoring inference details.}
We adopt the following procedure to assess the quality during evaluation:

\[
\mathcal{Q} = \sum_{i=1}^5 \omega_i \frac{e^{\mathcal{P}_{\textit{quality\_levels}[i]}}}{\sum_{i=1}^5 e^{\mathcal{P}_{\textit{quality\_levels}[i]}}},
\]
where \textit{quality\_levels} refers to a list of predefined quality levels: \textit{[High, Good, Fair, Poor, Low]}, and \(\mathcal{P}\) denotes the model's \textbf{logit outputs} for each quality level. Specifically, the vector corresponding to the quality description word in the model’s output sequence is first extracted, where its dimension matches the tokenizer's vocabulary size (located at the $-3$ index in our model). The logit values at the specific indices of this vector, which correspond to the $5$ quality level in the tokenizer’s vocabulary (indices $1550$, $1661$, $6624$, $7852$, and $3347$ in our model), are then selected. These logits are subsequently normalized using the softmax function. 

The values \(\omega\) represent the weight factors assigned to the normalized probabilities of each quality level, given by $[1, 0.75, 0.5, 0.25, 0]$. The resulting weighted sum of these probabilities produces the predicted quality score \(\mathcal{Q}\), which is confined within the range of $[0,1]$.

\paragraph{Quality interpreting inference details.}
For the quality understanding task, we use \textit{model.generate()} with \textit{greedy search} to ensure the reproducibility of the results. For multiple-choice questions in the benchmarks, we compare the first letter of the output (usually the selected option) with the correct answer and report the accuracy. For open-ended questions and multiple-choice questions where the first letter is not an option, we use \textit{GPT-5-nano} for judgment (except for \textit{GPT-4o} itself) since this is actually a textual analysis task with no need for powerful LMMs. For multiple-choice questions, we directly assess whether the answer is correct (scoring 0 or 1). For open-ended questions, we evaluate them based on three criteria: \textit{completeness}, \textit{accuracy}, and \textit{relevance}, with a score of 0, 1, or 2. The specific evaluation standards are as follows:

\textit{``Given the [question], evaluate whether the response [answer] completely matches the [correct answer]. 
First, check the response, and please rate the score 0 if the response is not a valid answer.
Please rate score 2 if the response completely or almost completely matches the correct answer on completeness, accuracy, and relevance. 
Please rate score 1 if the response partly matches the correct answer on completeness, accuracy, and relevance.
Please rate score 0 if the response doesn't match the correct answer on completeness, accuracy, and relevance at all.
Please provide the result in the following format: Score:"}

We set up $5$ rounds of \textit{GPT} scoring for each question. The final score for the question is determined by ``majority voting", selecting the most frequently occurring score. Based on our experiments, there has been no instance where the score distribution resulted in a ``2/2/1" split.

\subsection{Supplementary Experiments}
\label{Supplementary Experiments}
\paragraph{The Rationality on the Vision-Encoder-centered Setting}
We conduct experiments to rationalize the choice of vision-encoder-centered training. First, we replace the entire pretraining process with full-parameter fine-tuning, applying the same structural transfer operations to obtain the \textit{Zero} and \textit{Warm-up} series (\textit{Full-finetuning Reference}). We then compare these models with the corresponding series trained using the vision-encoder-centered approach (our setting in the main paper) across 15 datasets. The results are shown in Tabs.~\ref{tab:warmup} and \ref{tab:warm-up-IQA}. Experimental results show that during pretraining, fine-tuning only the vision encoder or performing full-parameter fine-tuning results in nearly \textbf{identical} performance on the scoring task (comparing the \textbf{Base} models). However, after full fine-tuning, the transfer performance of the corresponding \textit{Zero} and \textit{Warm-up} series exhibits a noticeable decline. This highlights that training centered around the vision encoder is crucial for the pretraining task, especially for scoring tasks.

\paragraph{Further Proof of Model Transferability}
We use the same warm-up data to fine-tune the base models (\textit{InternVL-1B, InternVL-2B, and InternVL-14B}) from scratch, which is recorded as \textit{warm-up (reference)}. The results are  also documented in Tabs.~\ref{tab:warmup} and \ref{tab:warm-up-IQA}. Additionally, we apply more heterogeneous LLM decoders (including \textit{Qwen2.5-7B}, \textit{Qwen2-7B}, and \textit{Internlm-2.5-7B}), without training on general LMMs to further test the model transferability, obtaining corresponding \textit{Zero} models (\textit{Additional VITAL-Zero Models}). The results are recorded in Tabs.~\ref{tab:warmup} and \ref{tab:warm-up-IQA}. Experimental results demonstrate that the \textit{VITAL-Warm-up} series shows significant advantages over the results obtained by fine-tuning the base model with the same warm-up data (the \textit{warm-up reference}), particularly in video quality scoring. This further emphasizes the importance of pre-training in enhancing model structural transferability and data efficiency. The additional \textit{VITAL-Zero} series has also shown good adaptability across a broader range of heterogeneous decoders (not only the \textit{Qwen2.5} series). This demonstrates the model structural transferability and adaptability gained from pretraining.

\paragraph{Linear-Probe Experiment Details}
We extract features from the 6th, 12th, and 18th layers (1024 dimensions each) of the image encoder (the \textit{InternViT}), and concatenate these features to form the image feature tokens. Next, we extract motion feature tokens (256 dimensions) from \textit{SlowFast}. All tokens were mapped to 3584 dimensions using their corresponding projectors, after which we averaged the tokens at each position to obtain four feature tokens (the 6th, 12th, 18th layer features from \textit{InternViT} and the \textit{SlowFast} features). These feature tokens were concatenated and passed through a linear layer without non-linear activation to produce logits for the five quality levels. \textit{CE loss} was used for training. During training, no validation was performed, and the model was trained for 5 full epochs using \textit{LSVQ (train)}. The trained model was then tested directly. For the \textit{Simple-VQA} reference model, we used its open-source weights (with \textit{Swin-B} and \textit{SlowFast} backbone) trained on \textit{LSVQ (train)} for testing.
\begin{table*}[t]\tiny
    \centering
    \renewcommand\arraystretch{1}
    \renewcommand\tabcolsep{2.5pt}
    \belowrulesep=0pt\aboverulesep=0pt

    \caption{Performance of \textit{VITAL-Seris} models and their reference counterparts (directly ``warm-up" from base models (denoted as \textit{Warm-up Reference}) or through full-parameter pretraining (denoted as \textit{Full-finetuning Reference})) on the video quality scoring task. Moreover, there are additional VITAL-Zero models (with other heterogeneous decoders). Datasets marked in \textit{italics} denote \textit{OOD} datasets. \textit{Mini-VQA} refers to \textit{Minimalistic-VQA}. }
    \vspace{-10pt}
    \resizebox{\linewidth}{!}{%
    \begin{tabular}{l|ccccccccccccccccc}
    \hline
    \multicolumn{1}{l|}{\textbf{Datasets}}
      & \multicolumn{2}{c}{\textbf{LSVQ(1080p)}}
      & \multicolumn{2}{c}{\textbf{LSVQ(test)}}
      & \multicolumn{2}{c}{\textbf{LIVE-VQC}}
      & \multicolumn{2}{c}{\textbf{KoNViD-1K}}
      & \multicolumn{2}{c}{\textbf{YT-UGC}}
      & \multicolumn{2}{c}{\textbf{\textit{YT-Gaming}}}
      & \multicolumn{2}{c}{\textbf{\textit{CGVDS}}}  & \multicolumn{2}{c}{\textbf{\textit{KVQ}}}&\multirow{3}{*}{\textbf{AVG.$\uparrow$}}\\
    \cline{1-17}
    \multicolumn{1}{l|}{\textbf{\# of videos}}
      & \multicolumn{2}{c}{\textbf{3,573}}
      & \multicolumn{2}{c}{\textbf{7,182}}
      & \multicolumn{2}{c}{\textbf{585}}
      & \multicolumn{2}{c}{\textbf{1,200}}
      & \multicolumn{2}{c}{\textbf{1,098}}
      & \multicolumn{2}{c}{\textbf{600}}
      & \multicolumn{2}{c}{\textbf{357}}  & \multicolumn{2}{c}{\textbf{2,926}}&\\
    \cline{1-17}
      \textbf{Models} & SRCC$\uparrow$ & PLCC$\uparrow$
      & SRCC$\uparrow$ & PLCC$\uparrow$
      & SRCC$\uparrow$ & PLCC$\uparrow$
      & SRCC$\uparrow$ & PLCC$\uparrow$
      & SRCC$\uparrow$ & PLCC$\uparrow$
      & SRCC$\uparrow$ & PLCC$\uparrow$ &SRCC$\uparrow$ & PLCC$\uparrow$&SRCC$\uparrow$ & PLCC$\uparrow$ &\\
    \cdashline{1-18}
    \multicolumn{1}{l}{\textit{Full-finetuning Reference}}\\ 
      \cdashline{1-18}
      \textsc{Base-8B-Full} &0.794&0.824&0.878&0.875&0.805&0.851&0.874&0.878&0.851&0.858&0.702&0.758&0.828&0.857&0.742&0.758&0.821\\
      \textsc{Warm-up-1B-Full} &0.722 & 0.750 & 0.805 & 0.802 & 0.712 & 0.758 & 0.802 & 0.806 & 0.754 & 0.756 & 0.675 & 0.722 & 0.755 & 0.769 & 0.681 & 0.693 & 0.748
\\
       \textsc{Warm-up-2B-full} &0.715 & 0.741 & 0.801 & 0.799 & 0.691 & 0.742 & 0.803 & 0.808 & 0.765 & 0.766 & 0.651 & 0.699 & 0.756 & 0.768 & 0.624 & 0.642 & 0.735
\\
        \textsc{Warm-up-14B-full} &0.715 & 0.747 & 0.811 & 0.803 & 0.690 & 0.758 & 0.813 & 0.814 & 0.755 & 0.753 & 0.646 & 0.677 & 0.754 & 0.765 & 0.659 & 0.686 & 0.740

      \\ 
      \textsc{Zero-1B-full} &0.611 & 0.540 & 0.639 & 0.645 & 0.641 & 0.646 & 0.657 & 0.592 & 0.579 & 0.563 & 0.469 & 0.406 & 0.594 & 0.645 & 0.402 & 0.303 & 0.558\\
       \textsc{Zero-2B-full} &0.594 & 0.559 & 0.698 & 0.686 & 0.589 & 0.641 & 0.760 & 0.659 & 0.609 & 0.599 & 0.508 & 0.529 & 0.640 & 0.607 & 0.493 & 0.501 & 0.604\\
        \textsc{Zero-14B-full} &0.600 & 0.594 & 0.760 & 0.672 & 0.670 & 0.626 & 0.740 & 0.628 & 0.737 & 0.675 & 0.509 & 0.572 & 0.725 & 0.721 & 0.478 & 0.463 & 0.636

      \\ 
      \cdashline{1-18}
    \multicolumn{1}{l}{\textit{Warm-up Reference}}\\ 
      \cdashline{1-18}
      \textsc{Warm-up-1B-Reference} &0.691&0.736&0.76&0.757&0.65&0.702&0.746&0.75&0.762&0.764&0.582&0.641&0.726&0.765&0.56&0.587&0.699\\
       \textsc{Warm-up-2B-Reference} &0.715&0.75&0.771&0.77&0.708&0.749&0.786&0.787&0.771&0.768&0.572&0.638&0.721&0.745&0.578&0.602&0.714\\
        \textsc{Warm-up-14B-Reference} &0.506&0.567&0.604&0.584&0.374&0.411&0.585&0.592&0.592&0.599&0.246&0.27&0.301&0.369&0.364&0.336&0.456
      \\ 
    \cdashline{1-18}
      \multicolumn{1}{l}{\textit{VITAL-Series}}\\ 
      \cdashline{1-18}

    \textsc{VITAL-Base-8B}  &0.786&0.815&0.883&0.879&0.800&0.843&0.878&0.881&0.854&0.856&0.710&0.764&0.830&0.854&0.721&0.766&0.820 \\
    \textsc{VITAL-Zero-1B}  &0.655&0.636&0.715&0.709&0.665&0.670&0.672&0.680&0.643&0.637&0.481&0.503&0.679&0.674&0.428&0.330&0.611
     \\ 
    \textsc{VITAL-Zero-2B} &0.660&0.686&0.765&0.750&0.688&0.713&0.783&0.747&0.676&0.655&0.582&0.615&0.759&0.693&0.562&0.595&0.683
       \\ 
       \textsc{VITAL-Zero-14B} &0.663&0.689&0.802&0.732&0.719&0.729&0.782&0.717&0.790&0.732&0.567&0.604&0.766&0.761&0.553&0.505&0.694 \\
    \textsc{VITAL--Warm-up-1B}
      & 0.786&0.819&0.869&0.866&0.771&0.812&0.873&0.875&0.846&0.848&0.705&0.754&0.808&0.824&0.730&0.743&0.808\\
    \textsc{VITAL-Warm-up-2B} 
    &0.787&0.817&0.869&0.866&0.768&0.816&0.874&0.878&0.843&0.844&0.700&0.750&0.807&0.819&0.654&0.669&0.798\\
    \textsc{VITAL-Warm-up-14B}
      & 0.787&0.806&0.870&0.865&0.773&0.819&0.875&0.877&0.843&0.841&0.688&0.722&0.804&0.815&0.740&0.765&0.795 \\
      \cdashline{1-18}
       \multicolumn{1}{l}{\textit{Additional VITAL-Zero Models}}\\ 
      \cdashline{1-18}
     \textsc{VITAL-Zero(Qwen2-7B)}
&0.699&0.730&0.784&0.764&0.677&0.701&0.812&0.805&0.465&0.538&0.542&0.565&0.538&0.529&0.337&0.332&0.614\\
\textsc{VITAL-Zero(Qwen2.5-7B)} 
  &0.781&0.793&0.864&0.841&0.718&0.772&0.867&0.845&0.826&0.817&0.708&0.75&0.8&0.797&0.627&0.667&0.780\\
\textsc{VITAL-Zero(InternLM-7B)} 
  &0.679 & 0.722 & 0.765 & 0.744 & 0.639 & 0.667 & 0.767 & 0.779 & 0.455 & 0.501 & 0.538 & 0.551 & 0.498 & 0.502 & 0.308 & 0.308 & 0.589

\\
    
    \cline{1-18}
    \end{tabular}%
    }
    \vspace{-1pt}
    \label{tab:warmup}
\end{table*}
\begin{table*}[tbp]
\vspace{-1em}
\centering
\small
\renewcommand\arraystretch{1}
\renewcommand\tabcolsep{7.5pt}
\caption{Performance of \textit{VITAL-Series} models and their reference counterparts (directly ``warm-up" from base model (denoted as \textit{Warm-up Reference}) or through full-parameter funetuning pretraining (denoted as \textit{Full-finetuning Reference})) on the image quality scoring task. Moreover, there are additional VITAL-Zero models (with other heterogeneous decoders).}
\vspace{-10pt}
\resizebox{\linewidth}{!}{%
\begin{tabular}{l|ccccccccccccccc}
\cline{1-16}
\textbf{Datasets}
  & \multicolumn{2}{c}{\textbf{KonIQ}}
  & \multicolumn{2}{c}{\textbf{SPAQ}}
  & \multicolumn{2}{c}{\textbf{LIVE-C}}
  & \multicolumn{2}{c}{\textbf{\textit{AGIQA}}}
  & \multicolumn{2}{c}{\textbf{\textit{KADID}}}& \multicolumn{2}{c}{\textbf{\textit{TID}}}&\multicolumn{2}{c}{\textbf{\textit{CSIQ}}}&\multirow{3}{*}{\textbf{AVG.$\uparrow$}}
 \\ 
\cline{1-15}
\textbf{\# of images}
  & \multicolumn{2}{c}{\textbf{2,010}}
  & \multicolumn{2}{c}{\textbf{2,224}}
  & \multicolumn{2}{c}{\textbf{1,169}}
  & \multicolumn{2}{c}{\textbf{2,982}}
  & \multicolumn{2}{c}{\textbf{2,000}}& \multicolumn{2}{c}{\textbf{3,000}}&\multicolumn{2}{c}{\textbf{866}}&
 \\ 
\cline{1-15}
  \textbf{Models}& SRCC$\uparrow$ & PLCC$\uparrow$ & SRCC$\uparrow$ & PLCC$\uparrow$ & SRCC$\uparrow$ & PLCC$\uparrow$ & SRCC$\uparrow$ & PLCC$\uparrow$ & SRCC$\uparrow$ & PLCC$\uparrow$ & SRCC$\uparrow$ & PLCC$\uparrow$ & SRCC$\uparrow$ & PLCC$\uparrow$ &  \\
\cdashline{1-16}
 \multicolumn{1}{l}{\textit{Full-finetuning Reference}}\\ 
      \cdashline{1-16}
      \textsc{Base-8B-Full} &0.928&0.933&0.872&0.875&0.855&0.871&0.740&0.815&0.773&0.719&0.652&0.69&0.817&0.838&0.813\\
      \textsc{Warm-up-1B-Full} &0.846 & 0.844 & 0.787 & 0.792 & 0.764 & 0.773 & 0.749 & 0.793 & 0.613 & 0.601 & 0.613 & 0.602 & 0.731 & 0.737 & 0.732\\
       \textsc{Warm-up-2B-full} &0.849 & 0.845 & 0.803 & 0.808 & 0.780 & 0.787 & 0.723 & 0.785 & 0.607 & 0.589 & 0.595 & 0.604 & 0.736 & 0.745 & 0.733
\\
        \textsc{Warm-up-14B-full} &0.846 & 0.844 & 0.807 & 0.813 & 0.791 & 0.789 & 0.683 & 0.750 & 0.627 & 0.605 & 0.598 & 0.612 & 0.721 & 0.730 & 0.730

      \\ 
      \textsc{Zero-1B-full} &0.725 & 0.675 & 0.695 & 0.693 & 0.590 & 0.560 & 0.594 & 0.616 & 0.579 & 0.590 & 0.611 & 0.602 & 0.602 & 0.605 & 0.624
\\
       \textsc{Zero-2B-full} &0.774 & 0.747 & 0.752 & 0.746 & 0.711 & 0.683 & 0.626 & 0.638 & 0.554 & 0.559 & 0.593 & 0.578 & 0.684 & 0.697 & 0.667
\\
        \textsc{Zero-14B-full} &0.786 & 0.788 & 0.780 & 0.756 & 0.719 & 0.688 & 0.626 & 0.635 & 0.624 & 0.592 & 0.542 & 0.577 & 0.692 & 0.688 & 0.678

      \\ 
      \cdashline{1-16}
      \multicolumn{1}{l}
      {\textit{Warm-up-Reference}}\\ 
\cdashline{1-16}

\textsc{Warm-up-1B-Reference}
  &0.828&0.832&0.844&0.842&0.69&0.751&0.73&0.811&0.632&0.647&0.679&0.703&0.676&0.763&0.745\\
\textsc{Warm-up-2B-Reference} 
  &0.875&0.87&0.856&0.847&0.76&0.802&0.757&0.827&0.621&0.631&0.665&0.681&0.713&0.789&0.764\\
\textsc{Warm-up-14B-Reference}
  &0.737&0.701&0.56&0.652&0.618&0.586&0.726&0.719&0.536&0.541&0.478&0.498&0.509&0.496&0.597\\
\cdashline{1-16}
\multicolumn{1}{l}{\textit{VITAL-Series}}\\ 
\cdashline{1-16}
 \textsc{VITAL-Base-8B}&0.931&0.931&0.884&0.886&0.851&0.866&0.736&0.811&0.759&0.708&0.680&0.707&0.823&0.851&0.816\\
 \textsc{VITAL-Zero-1B}&0.819&0.744&0.765&0.762&0.649&0.610&0.647&0.665&0.614&0.636&0.663&0.650&0.650&0.653&0.681\\
    \textsc{VITAL-Zero-2B} &0.897&0.855&0.862&0.856&0.810&0.783&0.725&0.738&0.625&0.630&0.662&0.648&0.764&0.781&0.760 \\
    \textsc{VITAL-Zero-14B} &0.876&0.878&0.868&0.841&0.800&0.765&0.709&0.721&0.706&0.665&0.610&0.652&0.774&0.765&0.759 \\
    \textsc{VITAL-Warm-up-1B}
      & 0.918&0.917&0.871&0.877&0.851&0.859&0.716&0.783&0.674&0.659&0.674&0.667&0.787&0.793&0.789 \\
    \textsc{VITAL-Warm-up-2B} &0.919&0.916&0.873&0.878&0.852&0.859&0.719&0.781&0.674&0.646&0.652&0.663&0.782&0.792&0.786\\
    \textsc{VITAL-Warm-up-14B}
      & 0.918&0.917&0.870&0.875&0.853&0.857&0.720&0.784&0.670&0.645&0.676&0.691&0.786&0.796&0.791\\
  \cdashline{1-16}
  \multicolumn{1}{l}
{\textit{Additional VITAL-Zero Models}}\\ 
\cdashline{1-16}
\textsc{VITAL-Zero(Qwen2-7B)}
&0.876&0.889&0.831&0.838&0.811&0.814&0.733&0.794&0.675&0.622&0.613&0.66&0.769&0.786&0.765\\
\textsc{VITAL-Zero(Qwen2.5-7B)} 
  &0.924&0.921&0.874&0.875&0.851&0.859&0.73&0.776&0.708&0.696&0.62&0.668&0.82&0.845&0.798\\
\textsc{VITAL-Zero(InternLM-7B)} 
  &0.870 & 0.879 & 0.825 & 0.804 & 0.792 & 0.786 & 0.687 & 0.778 & 0.652 & 0.583 & 0.598 & 0.652 & 0.751 & 0.774 & 0.745
\\
\cline{1-16}
\end{tabular}}
\vspace{-5pt}
\label{tab:warm-up-IQA}
\end{table*}
\begin{table*}[t]\tiny
    \centering
    \renewcommand\arraystretch{1.07}
    \renewcommand\tabcolsep{9pt}
    \belowrulesep=0pt\aboverulesep=0pt

    \caption{Evaluation results of \textit{VITAL-Assistant} and \textit{VQA\textsuperscript{2}-Assistant-Enhanced} on the \textit{Qbench-video-test-single}.
}
 \vspace{-10pt}
    \resizebox{\linewidth}{!}{\begin{tabular}{l|ccc|cccc|c}
  
       \hline
     \textbf{Sub-categories} & \multicolumn{3}{c|}{\textbf{Question Types}} & \multicolumn{4}{c|}{\textbf{Quality Concerns}} &\multirow{2}{*}{\textit{Overall$\uparrow$}}
        \\ \cline{1-8}
        \multirow{1}{*}{\textbf{Models}} & \textit{Binary} $\uparrow$&\textit{Multi.} $\uparrow$& \textit{Open-ended} $\uparrow$& \textit{Technical} $\uparrow$&\textit{Aesthetic} $\uparrow$& \textit{Temporal}  $\uparrow$&\textit{AIGC} $\uparrow$& \\
      \hdashline
      \textsc{VQA\textsuperscript{2}-Assistant-Enhanced} (the annotator)&69.70\% & 70.73\% & 40.66\% & 59.64\% & 58.53\% & 56.80\% & 55.59\% & 59.83\%\\ 
      \textsc{VITAL-Assistant-8B}&72.05\% & 72.13\% & 46.52\% & 61.75\%& 64.45\% & 61.56\% & 60.25\% & 63.11\%\\ 
     
      \cline{1-9}
      \end{tabular}}
      \vspace{-12pt}

    \label{tab:annotator_compare}
\end{table*}

\paragraph{Comparison with the \textit{VQA\textsuperscript{2}-Assistant-Enhanced}}
We compare the performance of \textit{VITAL-Assistant-8B} with the original in-domain annotator (\textit{VQA\textsuperscript{2}-Assistant-Enhanced}) for the quality interpreting task on \textit{QBench-Video-Test-Single} (see Tab.~\ref{tab:annotator_compare}). The experimental results show that \textit{VITAL-Assistant} outperforms its original annotator.

\vspace{-5pt}
\subsection{Justification on Annotator's Selection}
For the scoring task, the VQA and IQA models we selected include some of the most renowned SOTA models from recent years. These models vary in parameter size and design approaches, and importantly, they are all with open-source full weights and runnable code. While other models could have been selected, we believe the models chosen are the most representative, as others are either older or structurally or conceptually similar to the current models.

For the quality description task, \textit{VQA2-Assistant-Enhanced} is \textbf{the only} open-sourced in-domain LMM that possesses both image and video quality annotation capabilities, making the model selection reasonable.

\subsection{Data Overlapping Check}
Given that the pretraining process utilizes a vast amount of \textit{in-the-wild} image/video, it is crucial to ensure that there is no serious overlap between the training and test datasets. We compare the names of each image/video in the training dataset with those in the test dataset to ensure that there are no duplicate samples.

\subsection{Justification on PMOD Construction}
In the machine opinion list collection, we first apply the following formula to use a nonlinear regression method to map the predicted results from each machine annotation method to the ground truth values on the \textit{LSVQ (test)} (video) and \textit{KonIQ (test)} (image) datasets, respectively:
\begin{equation}
Q_m^{\prime}=\gamma_3 \operatorname{Sigmoid}\left(\gamma_1 Q_m+\gamma_2\right)+\gamma_4,
\end{equation}
where $Q_m$ denotes the annotated raw score of a single machine annotator and $Q_m^{\prime}$ represents the mapped value. 
This yields the mapping parameters $\gamma_1,\gamma_2,\gamma_3,\gamma_4$ for each machine scorer. Then, we apply the obtained mapping parameters to the machine opinions of our own dataset. Finally, for both images and videos (separated), we combine the scores obtained from all six methods and uniformly scale them to the $[0-100]$ range.

For \textit{pairwise PMOD} construction:
\begin{equation}
  p^{\textit{pred}}(\textit{\uppercase\expandafter{\romannumeral1}}>\textit{\uppercase\expandafter{\romannumeral2}})=\Phi\left(\frac{\mu_\textit{\uppercase\expandafter{\romannumeral1}}^{\textit{pred}}-\mu_\textit{\uppercase\expandafter{\romannumeral2}}^{\textit{pred}}}{\sqrt{\left(\sigma_\textit{\uppercase\expandafter{\romannumeral1}}^{\textit{pred}}\right)^2+\left(\sigma_\textit{\uppercase\expandafter{\romannumeral2}}^{\textit{pred}}\right)^2}}\right),  
\end{equation}
where $\Phi$ is the \textbf{\textit{cumulative distribution function of the standard Gaussian distribution}}, and the two variables $\mu_\textit{\uppercase\expandafter{\romannumeral1}}^{\textit{pred}}$ and $\mu_\textit{\uppercase\expandafter{\romannumeral2}}^{\textit{pred}}$ can be directly derived from the discrete probability distribution of the five quality levels in the quality token.

\subsection{Justification on Other Key Issues}

\textbf{Why is the ``quality-scoring" task primarily used to validate the pretraining effect?}  

In our experiments validating the generalization and scalability of the \textit{VITAL-Series}, we chose the ``quality-scoring" task as the primary focus because the scoring task data comprises a significant portion of the training data (analogous to the vanilla \textit{CLIP}, which also focuses on classification tasks, similar to its training task). Additionally, the probability distribution-based testing approach is well-suited for evaluating OOD data or the zero-shot scoring capability of the \textit{Zero-series} models. For other quality evaluation tasks, such as text generation, since pretraining does not involve complex instruction following and the inevitable overfitting from supervised fine-tuning (SFT), directly using the \textit{VITAL-Series} models in quality interpreting benchmark tests does not yield prominent results (though still outperforming the base model). Instead, we use data efficiency experiments to demonstrate the effectiveness of pretraining on the text generation task (in Fig.~\ref{fig:plots} in the main paper). We believe large-scale pre-training significantly enhances the model's upper bound for visual quality evaluation, but \textbf{justifying this enhancement requires designing appropriate testing methods}, such as the quality scoring task used here. For other visual quality evaluation tasks, we believe specialized downstream task training or structural transfer is required to fully demonstrate the effect of pretraining.

\textbf{Why use the \textit{InternVL-Instruct} as the pretraining base rather than training from scratch?}  

The \textit{InternVL-Instruct} already possesses extensive visual question answering priors, including foundational knowledge for visual quality assessment tasks. This allows us to quickly transfer model functionality without the need to ``teach the model how to speak". Furthermore, the \textit{VITAL-Zero} series benefits from the shared vision-language priors in general LLM decoders, such as \textit{InternVL}, which is essential for enhancing the model's structural transferability. Using a general LMM as a base model is therefore both reasonable and necessary.

\textbf{Where does the performance advantage of \textit{VITAL-Series} in the quality scoring task lie?}

The performance advantage of the \textit{VITAL-Series} in the quality scoring task is primarily demonstrated in its more accurate scoring ability for OOD visual content types. For in-domain datasets like \textit{LSVQ (test)}, \textit{KoNViD-1K}, and \textit{SPAQ} (which primarily consist of in-the-wild UGC visual content and authentic distortions), the model performance has nearly saturated, and the differences between different models are minimal. However, for OOD data (including synthetic distortion content, various PGC content such as CG, AIGC, HD, HFR, etc.), the \textit{VITAL-Series} models demonstrate a clear advantage. We believe ensuring strong generalization performance is critical in current visual quality assessment tasks.

\textbf{Why is there no strict feature-based filtering for the candidate pool data selection?}  

Our goal was to \textbf{achieve data expansion without relying on prior feature-statistic knowledge for filtering} (e.g., sampling according to widely used datasets based on statistical feature distributions). This approach ensures the successful application of large-scale in-the-wild data. While specific prior standards may improve performance on certain test sets, they can potentially harm the model's \textbf{real-world generalization} ability.

\textbf{Why not use synthetic distortion data for the scoring task, and instead only include it in the text generation task?} 

Although synthetic distortion data is easy to generate, datasets like \textit{KADID}, \textit{TID}, and \textit{CSIQ}  are based on synthetic distortions, and including synthetic data in the scoring task could compromise the fairness of comparative experiments. Therefore, we avoided including synthetic data in the scoring task.

\textbf{Why are the image and motion projector modules included in the decoder part rather than the \textit{VITAL vision encoder}?}  

The reason for including the image and motion projectors in the decoder part rather than the vision encoder is that different sizes of LLMs correspond to different token hidden sizes (see Tab \ref{tab:hyperparam}). This mismatch means that unfreezing the projectors (typically used for dimensional adaptation) during pretraining would result in incompatibility with LLM decoders of different parameter sizes, making direct adaptation difficult.

\textbf{Why is the warm-up fine-tuning designed as decoder-only?}  
The design of the \textit{VITAL-Vision-Encoder} aims for zero-shot transfer to other heterogeneous decoder structures. Therefore, during the transfer process, the \textit{VITAL-Vision-Encoder} itself remains unchanged.

\textbf{Why use \textit{QBench-Video} as the evaluation set for the text generation task instead of the earlier \textit{QBench}?}  
\textit{QBench-Video} is a comprehensive evaluation set for video visual quality, which includes spatial visual quality for images and temporal-related quality for video features. It contains the two text generation pretraining subtasks: distortion identification and quality interpretation-related test questions. This makes it a more suitable benchmark for evaluating text generation tasks. In contrast, the \textit{QBench} series, which focuses on images and was released at a much earlier date, is too simple in structure and does not provide a satisfactory level of complexity. As a result, we exclude this earlier benchmark from our evaluation.

\section{Additional Dataset Materials}
\begin{table*}[h]
    \centering
    \caption{Summary of existing VQA MIDBs.}
    \vspace{-7pt}
    \small
    \renewcommand\arraystretch{0.9}
    \renewcommand\tabcolsep{0.5pt}
    \begin{tabular}{l|c|c|c|c|c|c}
        \hline
        \textit{MIDBs for VQA} &\# Videos & MOS &\# MOS &VL-pair &\# VL-pair& Description \\ 
        \hline
        LIVE-VQA& 160 & \ding{51} &160 & \ding{55} & / &Full-reference video quality rating  \\ 
        CVD2014&234 & \ding{51} &234 & \ding{55} & / & Quality assessment of video captured by cameras \\ 
        LIVE-Qualcomm&208 & \ding{51} &208 & \ding{55} & / & Mobile in-capture video quality rating \\ 
        KoNViD-1K& 1,200 & \ding{51} &1,200 & \ding{55} & /  & Unified UGC video quality rating \\ 
        LIVE-VQC& 585 & \ding{51} & 585 & \ding{55} & / & Quality rating of real world UGC videos \\ 
        YouTube-UGC& 1,380 & \ding{51} & 1,380  & \ding{55} & /& Quality rating of UGC videos \\ 
        LSVQ&39,075 & \ding{51} &39,075  & \ding{55} & / &Large-scale quality rating of UGC videos \\ 
        LIVE-NFLX-I&558 & \ding{51} & 558  & \ding{55} & / & Quality-of-experience (QoE) rating of hand-craft streaming videos \\ 
        LIVE-NFLX-II&420 & \ding{51} & 420  & \ding{55} & / & QoE rating of real-world streaming videos\\ 
        WaterlooSQoE-III& 450 & \ding{51} & 450  & \ding{55} & / & QoE rating of hand-craft streaming videos\\ 
        LBVD& 1,013 & \ding{51} & 1,013  & \ding{55} & / & QoE assessment of in-the-wild streaming videos \\ 
        WaterlooSQoE-IV& 1,350 & \ding{51} & 1,350  & \ding{55} & / & Large-scale QoE assessment of hand-craft streaming videos \\ 
        TaoLive& 3,762 & \ding{51} &3,762  & \ding{55} & / &Quality rating of live streaming (compresqsed) videos \\ 
        Maxwell& 4,543 & \ding{51} &9,086  & \ding{55} & / &Fine-grained (technical/aesthetic) quality rating of UGC videos \\ 
        VQA\textsuperscript{2}-Stage-1& 12,385 & \ding{55} & /  & \ding{51} &12,385 &Pre-training MIDB for distortion recognition \\ 
        VQA\textsuperscript{2}-Stage-2& 30,156 & \ding{51} & 30,156  & \ding{51} &30,156 &Large-scale MIDB specially for video quality rating. \\
        VQA\textsuperscript{2}-Stage-3& 15,500 & \ding{55} & /  & \ding{51} &115,214 &Human-annotated MIDB for video quality understanding. \\
        OmniVQA-Chat-20K& 20,000 & \ding{51} &20,000  & \ding{51} &20,000 &Large-scale MIDB for quality rating for in-the-wild UGC videos \\ 
        OmniVQA-MOS-400K& 86,716 & \ding{55} & /  & \ding{51} & 402,987 &Machine-vision-dominated MIDB for video quality understanding \\ 
        \hline
    \end{tabular}
    \vspace{-6pt}
    \label{tab:VQA_summary}
\end{table*}

\begin{table*}[h]
    \centering
    \caption{Summary of existing IQA MIDBs.}
    \vspace{-7pt}
    \small
    \renewcommand\arraystretch{0.9}
    \renewcommand\tabcolsep{0.5pt}
    \begin{tabular}{l|c|c|c|c|c|c}
        \hline
        \textit{MIDBs for IQA} &\# Images & MOS &\# MOS &VL-pair &\# VL-pair& Description \\ 
        \hline
        TID2013& 3000 & \ding{51} &3000 & \ding{55} & / &Full-reference, synthetic distortion  \\ 
        KonIQ-10K&10,073 & \ding{51} &10,073 & \ding{55} & / & No-reference, authentic distortion \\ 
        KADID-10K&10,125 & \ding{51} &10,125 & \ding{55} & / & No-reference, synthetic distortion, weakly-supervised \\ 
        SPAQ& 11,125 & \ding{51} &11,125 & \ding{55} & /  & No-reference, authentic distortion,smartphone photos \\ 
        AGIQA-3K& 2,982 & \ding{51} & 2,982 & \ding{55} & / & AIGC Images \\ 
        LIVE-C& 1,169 & \ding{51} & 1,169  & \ding{55} & /& Authentic distortions, in-the-wild images, no-reference. \\ 
        CSIQ&866 & \ding{51} &866& \ding{55} & / &Full-reference, synthetic distortion with DMOS. \\ 
        Q-Align-DB&15,800 & \ding{51} & 15,800  & \ding{51} & 15,800 & Reformulating existing IQA datasets (KonIQ+SPAQ) \\ 
        Q-Pathway-200K& 18,973 & \ding{55} & / & \ding{51} &200K &Human annotated image technical quality interpreting VL pairs.\\ 
        AesMMIT& 21,904 & \ding{55} &/  & \ding{51} & 409K &Human annotated image aesthetic quality interpreting VL pairs.\\ 
        DepictQA-V1& 197K & \ding{55} & 120,500  & \ding{51} & / & Human-in-the-loop annotated image technical quality.  \\ 
        DepictQA-V2& 140K & \ding{55} & 495K  & \ding{51} & / & Human-in-the-loop annotated(mainly with synthetic distortions).  \\ 
        \hline
    \end{tabular}
    \vspace{-6pt}
    \label{tab:IQA_summary}
\end{table*}

\subsection{Summary of Existing MIDBS}
We conduct a summary of existing MIDBs for VQA and IQA in Tabs.~\ref{tab:VQA_summary} and \ref{tab:IQA_summary}. This clearly demonstrates the scale advantage of our dataset.
\subsection{Statistic Information}
\begin{figure*}[t]
    \centering
    \includegraphics[width=\linewidth]{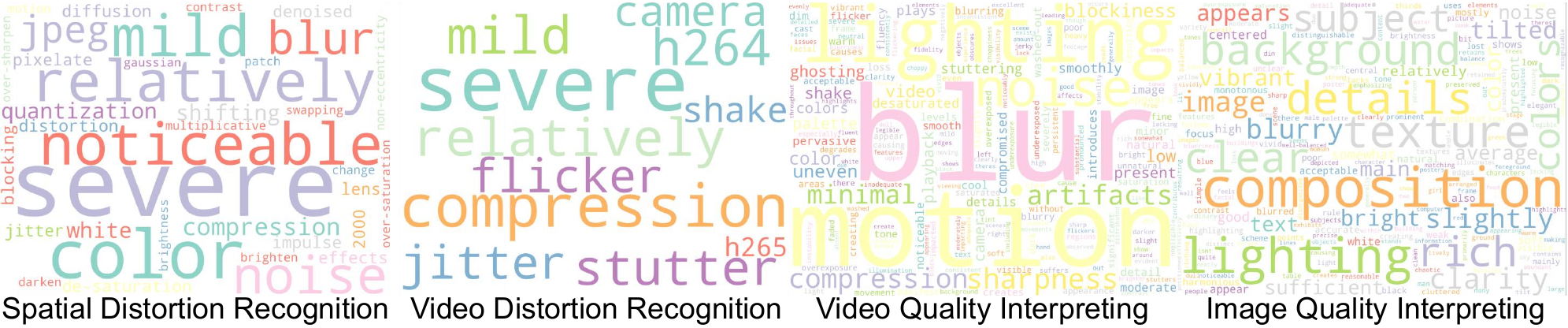}
    \vspace{-7pt}
    \caption{Wordclouds of the VL pairs in the text generation task.}
    \label{fig:wordcloud}   
\end{figure*}

\begin{figure*}[t]
    \centering
    \includegraphics[width=\linewidth]{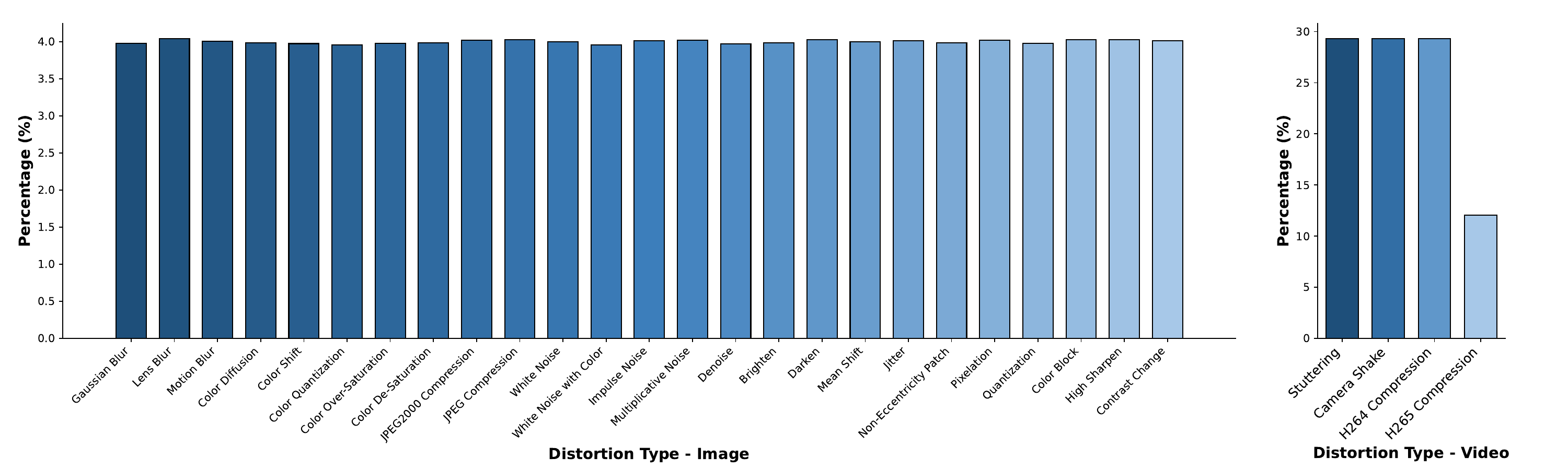}
    \vspace{-12pt}
    \caption{Statistics of distortion types.}
    \label{fig:number}   
\end{figure*}

\begin{figure*}[t]
    \centering
    \includegraphics[width=\linewidth]{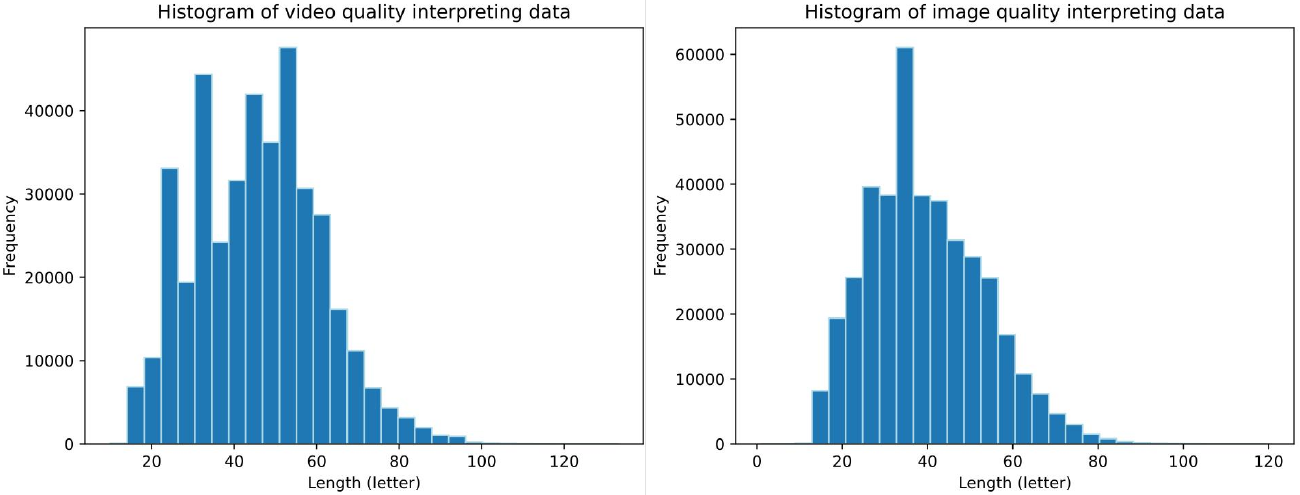}
    \vspace{-18pt}
    \caption{Sentence lengths of the VL pairs in the quality interpreting subtask.}
    \label{fig:lengths}   
\end{figure*}

\begin{figure*}[t]
    \centering
    \includegraphics[width=\linewidth]{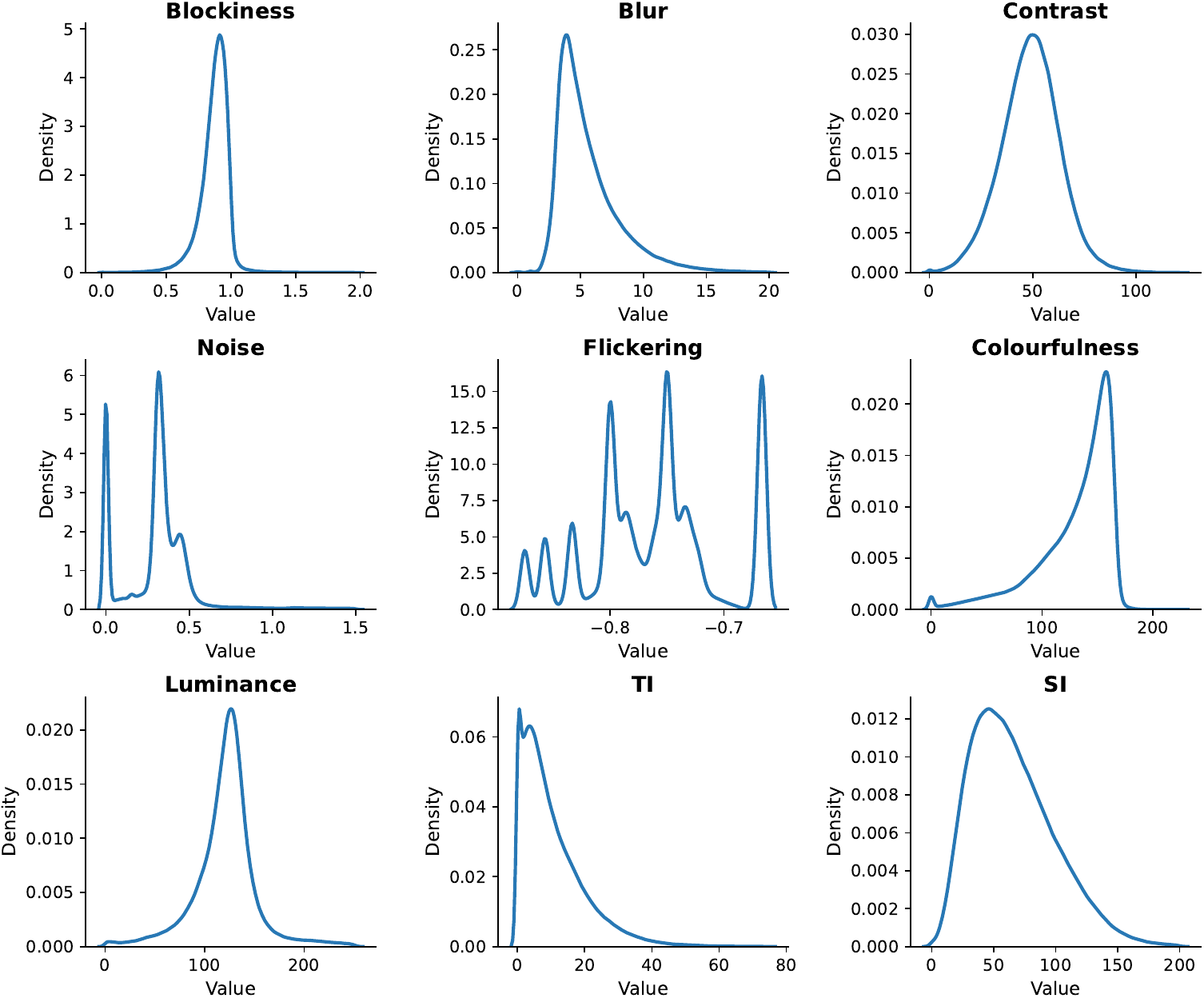}
    \vspace{-10pt}
    \caption{Statistics metrics for the video data in the pre-training dataset.}
    \label{fig:metrics}   
\end{figure*}

The word clouds of the VL pairs in the text generation task are shown in Fig. \ref{fig:wordcloud}. The proportion of samples for each distortion type of images and videos is illustrated in Fig.~\ref{fig:number}. The sentence lengths of the VL pairs in the quality interpreting subtask are shown in Fig.~\ref{fig:lengths}. The attribute metrics for the video data in the pre-training dataset are shown in Fig.~\ref{fig:metrics}.
\subsection{Dataset Construction Details}
\label{Dataset Construction}
For image data, we build our dataset through a large-scale web-scraping pipeline applied to \textit{Baidu Image Search}. To achieve broad coverage of real-world visual scenes, we construct a query list containing more than 30 high-level keywords—such as \textit{animals}, \textit{landscapes}, \textit{transportation}, etc. Each keyword is further expanded into multiple subtopics, enabling the crawler to retrieve diverse and representative visual content across various domains. All scraping procedures strictly follow ethical and legal constraints: only publicly available images with permissive usage licenses are collected, and no content requiring authentication or explicit authorization is included. 

For each video, we compute nine low-level quality metrics that capture fundamental visual characteristics. As illustrated in Fig.~\ref{fig:metrics}, the distributions of these metrics demonstrate that our dataset spans a wide and diverse range of visual quality conditions. We briefly interpret each metric as follows.
\textit{Blockiness}~\citep{romaniak2012perceptual} refers to blocking artifacts caused by aggressive compression, which is assessed by comparing luminance differences within and across block boundaries. 
\textit{Blur}~\citep{narvekar2011no} quantifies perceptual sharpness using the Cumulative Probability of Blur Detection, which evaluates edge-width statistics relative to human blur sensitivity. 
\textit{Contrast}~\citep{peli1990contrast} measures the dispersion of pixel intensities and indicates the overall dynamic range of the frame. 
\textit{Noise} captures random high-frequency fluctuations by estimating the residual energy remaining after low-pass filtering. 
\textit{Flickering}~\citep{pandel2008measuring} denotes unstable temporal variations identified by counting significant frame-to-frame luminance changes. 
\textit{Colourfulness}~\citep{hasler2003measuring} evaluates the distribution and balance of chromatic components based on opponent-color statistics. 
\textit{Luminance} measures the overall brightness level of each frame. 
\textit{Spatial information (SI)}~\citep{recommendation910subjective} characterizes spatial complexity using the standard deviation of Sobel-filtered edge responses. 
\textit{Temporal information (TI)}~\citep{recommendation910subjective} captures motion intensity by evaluating the variability of inter-frame differences.

We select $25$ types of spatial distortions (following the methodology in \textit{KADIS-700K}~\cite{lin2019kadid}) with five severity levels and four types of video-specific distortions with three severity levels. 
Here we provide a brief description of each \textbf{spatial distortion} type:

\noindent\textbf{Gaussian Blur.}
Gaussian blur smooths local image structures by convolving the image with a Gaussian kernel of standard deviation \( \sigma \):
\[
\tilde{I} = I * G_{\sigma}.
\]
As \( \sigma \) increases, edges become softer and fine textures gradually disappear, producing a natural smoothing effect.

\noindent\textbf{Lens Blur.}
Lens blur simulates optical defocus by filtering the image with a circular point spread function (PSF), represented by a normalized disk kernel of radius \( r \):
\[
\tilde{I} = I * D_{r}.
\]
This distortion produces blurred highlights and rounded defocus patterns similar to bokeh observed in real camera lenses.

\noindent\textbf{Motion Blur.}
Motion blur imitates camera or object motion during exposure by convolving the image with a linear PSF of length \( r \) and direction \( \theta \):
\[
\tilde{I} = I * M_{r,\theta}.
\]
The resulting image exhibits directional streaks whose strength depends on the blur length and orientation.

\noindent\textbf{Color Diffusion.}
Color diffusion selectively smooths chromatic information while preserving luminance. In Lab space, only the \( a \) and \( b \) channels are blurred:
\[
\tilde{a} = a * G_{\sigma},\qquad 
\tilde{b} = b * G_{\sigma}.
\]
This produces a soft watercolor-like appearance with reduced color contrast but intact structural edges.

\noindent\textbf{Color Shift.}
Color shift introduces subtle chromatic misalignment by spatially translating a single color channel (typically green) by a displacement \( \Delta \). Using a spatial blending mask \( M \), the shifted channel becomes:
\[
\tilde{G} = G(\cdot - \Delta)\, M + G(1 - M).
\]
This produces mild color fringing reminiscent of chromatic aberration or sensor alignment errors.

\noindent\textbf{Color Quantization.}
Color quantization compresses color richness by mapping each pixel to one of \( K \) representative palette entries:
\[
\tilde{I}(x,y) = \mathcal{C}[X(x,y)],
\]
where \( X \) is an indexed image and \( \mathcal{C} \) is the learned palette. Reducing \( K \) introduces banding artifacts and coarse color transitions.

\noindent\textbf{Color Over-Saturation.}
Over-saturation increases chromatic intensity by scaling the saturation channel in HSV space:
\[
\tilde{S} = \gamma S,\quad \gamma > 1.
\]
This results in vivid, exaggerated colors commonly seen in overly processed photographs.

\noindent\textbf{Color De-Saturation.}
De-saturation reduces color strength by attenuating the chromatic channels in Lab space:
\[
\tilde{a} = \beta a,\qquad 
\tilde{b} = \beta b,\quad 0 < \beta < 1.
\]
The output appears washed-out or faded, with weakened chromatic contrast.

\noindent\textbf{JPEG2000 Compression.}
JPEG2000 distortion arises from wavelet-based compression. Applying a compression–decompression cycle yields:
\[
\tilde{I} = \mathrm{Decode}\big(\mathrm{Encode}_{\mathrm{JP2}}(I)\big).
\]
Artifacts include edge ringing, wavelet granularity, and smooth but degraded textures.

\noindent\textbf{JPEG Compression.}
JPEG compression produces DCT-based artifacts after quantizing blockwise frequency coefficients:
\[
\tilde{I} = \mathrm{Decode}\big(\mathrm{Encode}_{\mathrm{JPEG}}(I)\big).
\]
Depending on the quality level, the result contains blocking, ringing, and loss of fine detail.

\noindent\textbf{White Noise.}
White noise adds pixelwise Gaussian noise to each color channel:
\[
\tilde{I} = I + \eta,\qquad 
\eta \sim \mathcal{N}(0,\sigma^2).
\]
This simulates sensor noise or low-light imaging degradation.

\noindent\textbf{White Noise with Color.}
This variant injects Gaussian noise into YCbCr channels before converting back to RGB:
\[
(Y',Cb',Cr') = (Y,Cb,Cr) + (\eta_Y,\eta_{Cb},\eta_{Cr}).
\]
Because chrominance is also corrupted, the noise exhibits visible color tints.

\noindent\textbf{Impulse Noise.}
Impulse noise replaces random pixels with extreme values:
\[
\tilde{I}(x,y)=
\begin{cases}
0 & \text{with probability }  p_i/2,\\
1 & \text{with probability }  p_i/2,\\
I(x,y) & \text{otherwise}.
\end{cases}
\]
This produces classic salt-and-pepper artifacts.

\noindent\textbf{Multiplicative Noise.}
Multiplicative (speckle) noise modulates intensity by a multiplicative Gaussian factor:
\[
\tilde{I} = I(1+\epsilon),\qquad 
\epsilon \sim \mathcal{N}(0,\sigma^2).
\]
This results in grainy, signal-dependent variations typical in coherent imaging systems.

\noindent\textbf{Denoise.}
The image is first corrupted with Gaussian noise and then passed through a CNN denoiser \( \mathcal{D} \):
\[
\tilde{I} = \mathcal{D}(I+\eta).
\]
Residual smoothing and faint artifacts remain due to imperfect restoration.

\noindent\textbf{Brighten.}
Brightening enhances mid-tone luminance using a nonlinear tone curve:
\[
\tilde{L} = \Gamma(L;\alpha_b).
\]
This increases brightness while maintaining overall dynamic range.

\noindent\textbf{Darken.}
Darkening applies the inverse tone curve:
\[
\tilde{L} = \Gamma(L;\alpha_d).
\]
It reduces luminance in the mid-range while preserving highlights and shadows.

\noindent\textbf{Mean Shift.}
Mean shift adds a constant offset to all pixels:
\[
\tilde{I} = \mathrm{clip}(I+\delta_m).
\]
This produces a uniform global change in brightness.

\noindent\textbf{Jitter.}
Jitter displaces pixels by random offsets and resamples the image:
\[
\tilde{I}(x,y)=I(x+\Delta_x,\ y+\Delta_y).
\]
The result exhibits irregular spatial jittering and local distortions.

\noindent\textbf{Non-Eccentricity Patch.}
Small patches are relocated to nearby positions to create local inconsistencies:
\[
\tilde{I}(\Omega'_k)=I(\Omega_k).
\]
Although local geometry is altered, global structure remains mostly preserved.

\noindent\textbf{Pixelation.}
Pixelation downscales the image and then upsamples it via nearest-neighbor interpolation:
\[
\tilde{I} = \mathrm{resize}_{nn}(I_{\downarrow}).
\]
This produces blocky, coarse patterns indicative of low-resolution displays.

\noindent\textbf{Quantization.}
Pixel intensities are mapped to discrete levels using multi-threshold partitioning:
\[
\tilde{I}(x,y)=q_j \quad \text{if } I(x,y)\in[t_j,t_{j+1}).
\]
This introduces tonal banding in smooth gradient regions.

\noindent\textbf{Color Block.}
Random solid-color rectangles are inserted into the image:
\[
\tilde{I}(x,y)=v_k \quad \text{for } (x,y)\in\Omega_k.
\]
This simulates occlusion or block-based corruption.

\noindent\textbf{High Sharpen.}
High sharpening applies an intensified unsharp mask:
\[
\tilde{I}=\mathrm{clip}(I+\alpha(I-B)).
\]
Large \( \alpha \) produces pronounced ringing and edge overshoots.

\noindent\textbf{Contrast Change.}
Contrast is modified using a nonlinear tone curve:
\[
\tilde{p}=\Gamma(p;\beta).
\]
Increasing or decreasing \( \beta \) adjusts the steepness of mid-tone transitions, altering global image contrast.

Here we provide a brief description of each \textbf{video-specific distortion} type:

\noindent\textbf{Stuttering.}
Stuttering simulates temporal freezing in videos by intermittently dropping and repeating frames. The distortion level determines the probability $q_s$ that the current frame is substituted by a previous one. For each frame $F_t$, a random variable $u \sim \mathcal{U}(0,1)$ is drawn, and the output frame is computed as
\[
\tilde{F}_t =
\begin{cases}
F_{t-1}, & \text{if } u \le q_s, \\
F_t, & \text{otherwise}.
\end{cases}
\]
This mechanism introduces irregular temporal progression and discontinuities, producing a visual effect similar to playback stuttering in low-quality or unstable video streams.

\noindent\textbf{Camera Shake.}
Camera shake simulates unintended hand-held camera motion by applying small random spatial perturbations to consecutive frames, with the shake intensity controlling the displacement magnitude. For each frame, slight horizontal and vertical offsets are introduced via an affine transformation:
\[
T =
\begin{bmatrix}
1 & 0 & \Delta_x \\
0 & 1 & \Delta_y
\end{bmatrix},
\]
where $\Delta_x$ and $\Delta_y$ are random shifts sampled according to the shake level. This produces a  frame-wise jitter characteristic of unstable hand-held video capture.

\noindent\textbf{H264 Compression.}
H264 compression applies the widely used AVC encoding standard to simulate streaming-related video degradation. Each video is encoded using three constant rate factor (CRF) values—24, 36, and 48—via the \texttt{fast} preset in \texttt{ffmpeg}. This configuration introduces typical AVC artifacts, including blocking, blurring, texture loss, and reduced edge fidelity.

\noindent\textbf{H265 Compression.}
H265 compression employs the HEVC standard to generate distortions representative of modern high-efficiency codecs. We apply the same CRF values (24, 36, and 48) as in H264 compression and encode all videos using the \texttt{very slow} preset in \texttt{ffmpeg}. This setting produces characteristic HEVC artifacts such as smoother textures, finer detail suppression, and mild blocking or ringing effects.

\section{Limitations}

As shown in Fig. \ref{fig:plots} in the main paper, the data-scaling effects from both pretraining and post-training have not yet reached the peak, and there remains room for exploration. However, due to resource limitations, we are unable to present the results of larger-scale data training in this work. Nevertheless, this is easily extendable in future studies and is one of the primary goals of our upcoming research.

\section{Applications}
\subsection{Potential Future Applications}
The \textit{VITAL-Vision-Encoder} can be applied to downstream tasks in more specialized visual quality evaluation scenarios, while the \textit{VITAL-Zero} (including linear probe) and \textit{VITAL-Warm-up} series can be applied to specific scenarios with varying model parameter size requirements.
\subsection{Case Studies}
\begin{figure*}[t]
    \centering
    \includegraphics[width=\linewidth]{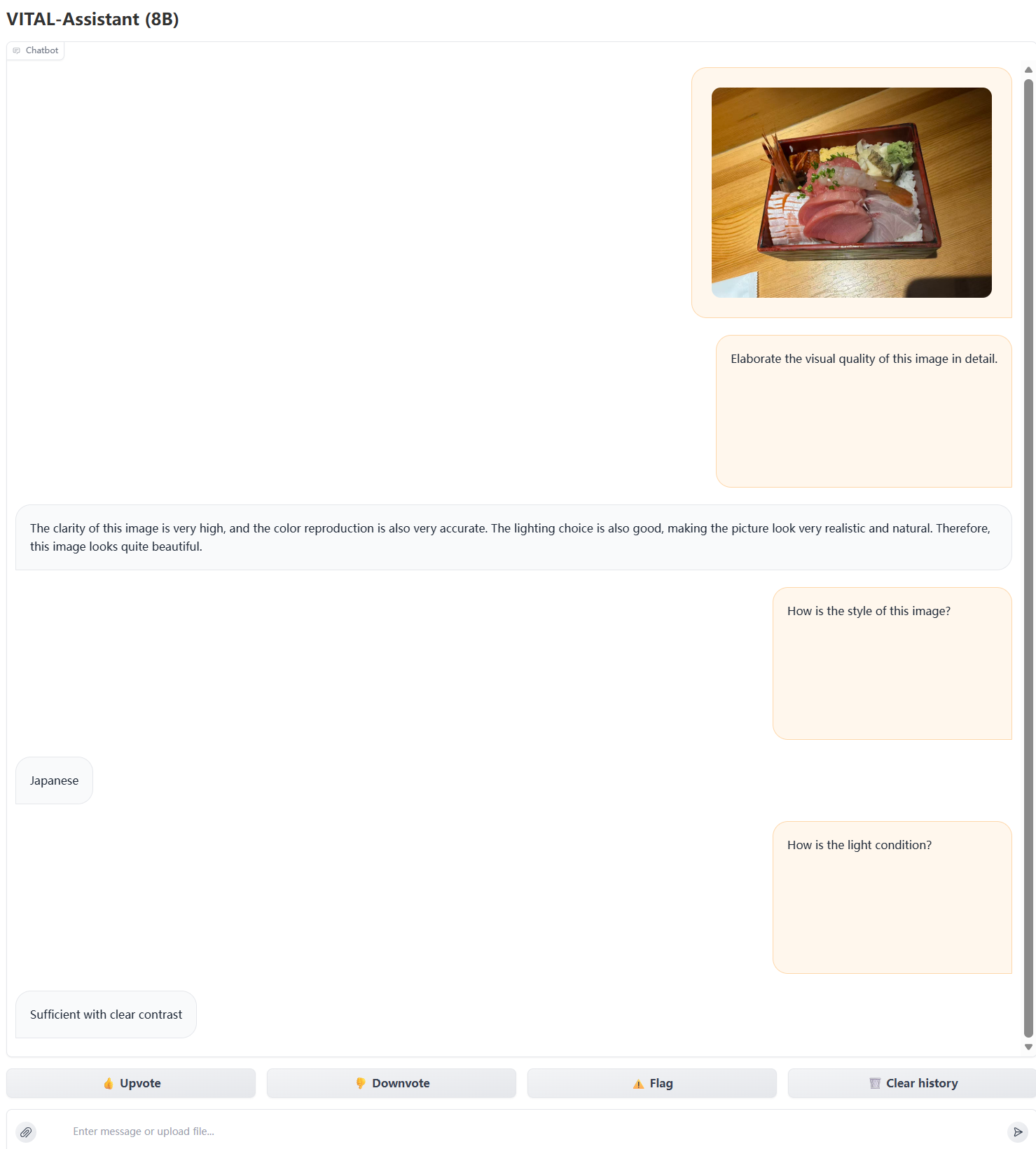}
    \vspace{-7pt}
    \caption{A photo taken by a smartphone in a Japanese restaurant.}
    \label{fig:case1}   
\end{figure*}

\begin{figure*}[t]
    \centering
    \includegraphics[width=\linewidth]{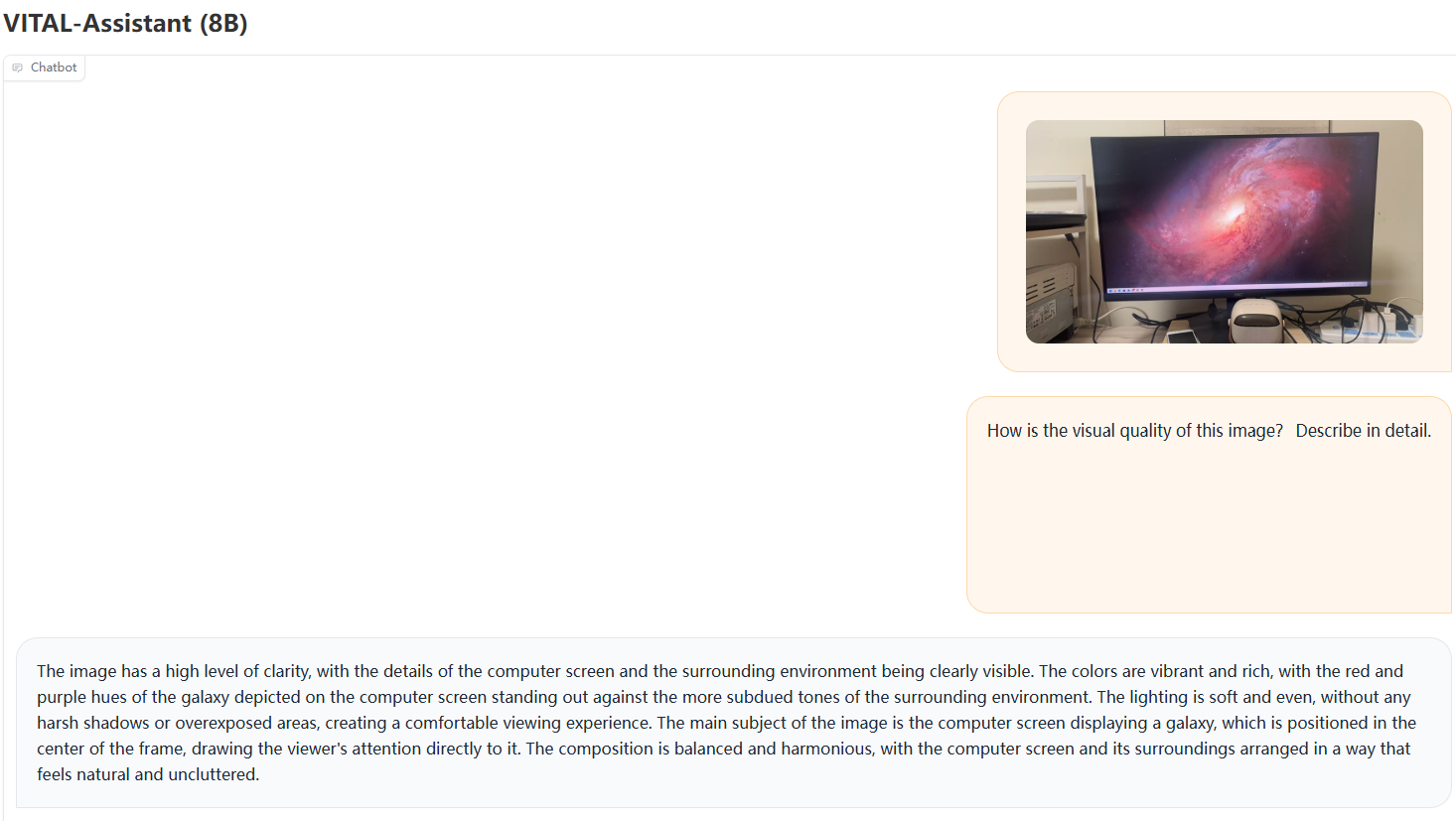}
    \vspace{-7pt}
    \caption{A photo taken by a camera in an office.}
    \label{fig:case2}   
\end{figure*}

\begin{figure*}[t]
    \centering
    \includegraphics[width=0.9\linewidth]{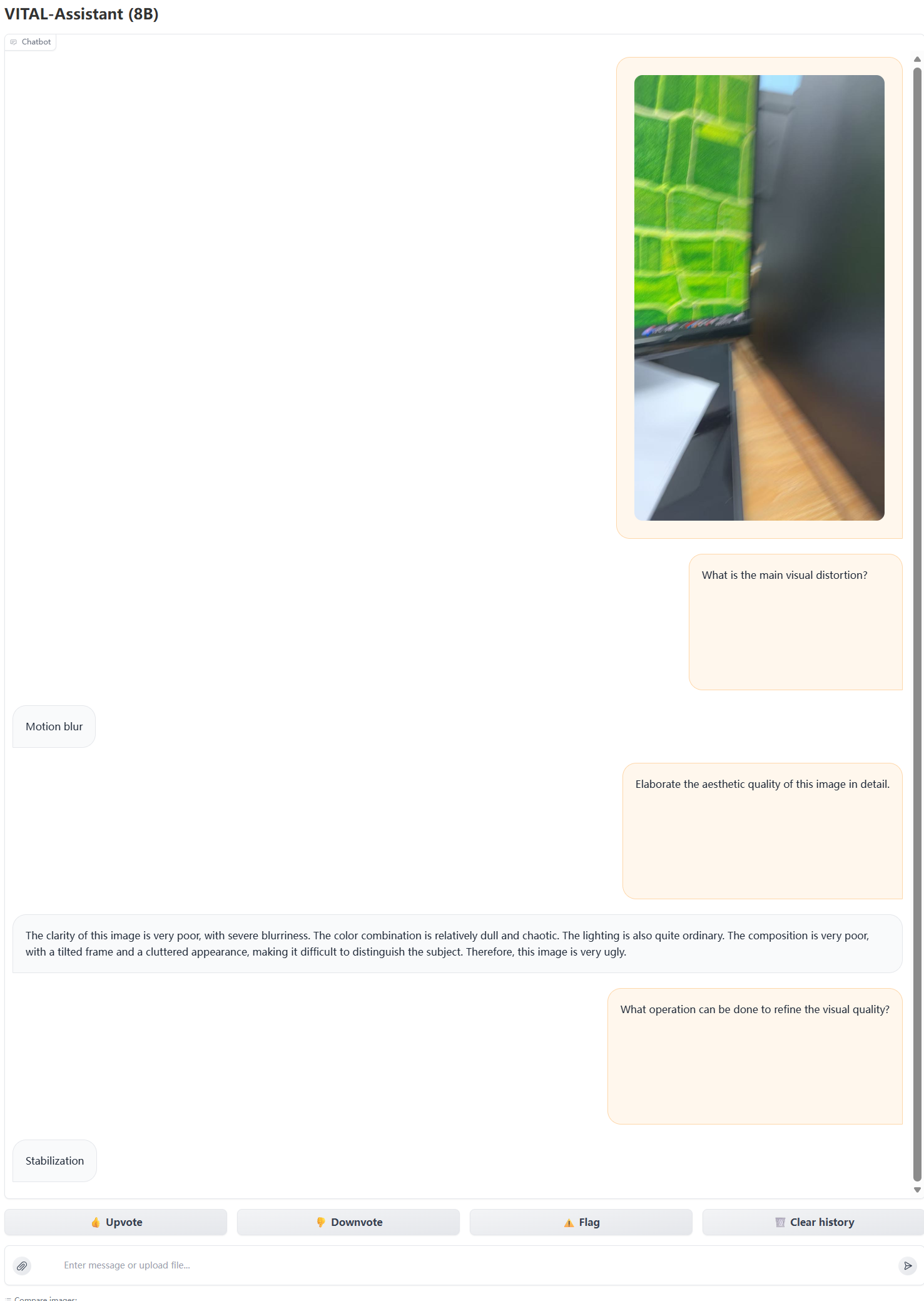}
    \vspace{-7pt}
    \caption{A photo taken by a smartphone, but without the necessary control of the visual quality.}
    \label{fig:case3}   
\end{figure*}

\begin{figure*}[t]
    \centering
    \includegraphics[width=0.8\linewidth]{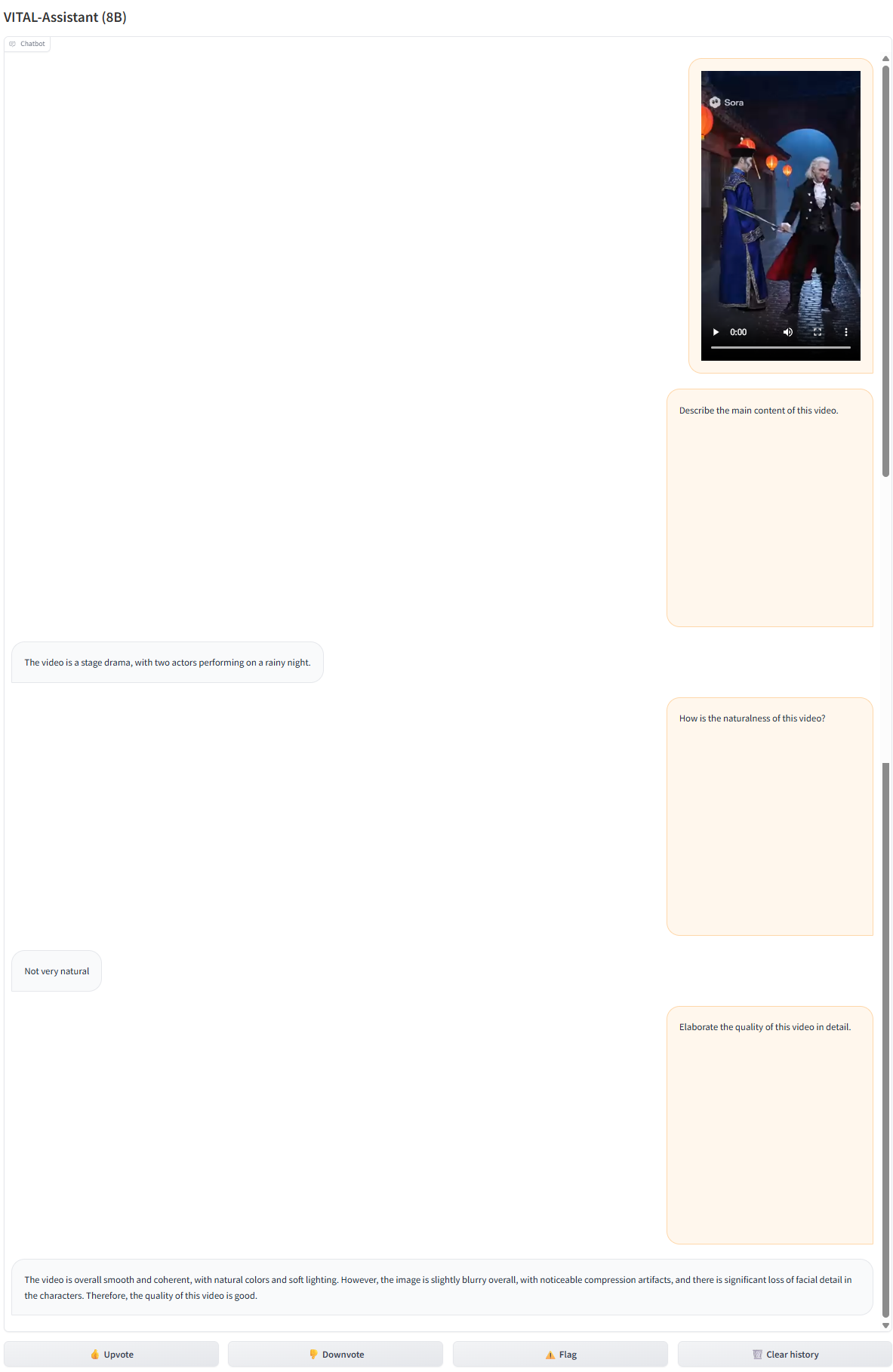}
    \vspace{-7pt}
    \caption{A video generated by \textit{Sora2} with the prompt: Generate a video of two actors, one playing a Chinese ancient zombie and the other playing a European vampire, battling each other.}
    \label{fig:case4}   
\end{figure*}

\begin{figure*}[t]
    \centering
    \includegraphics[width=0.9\linewidth]{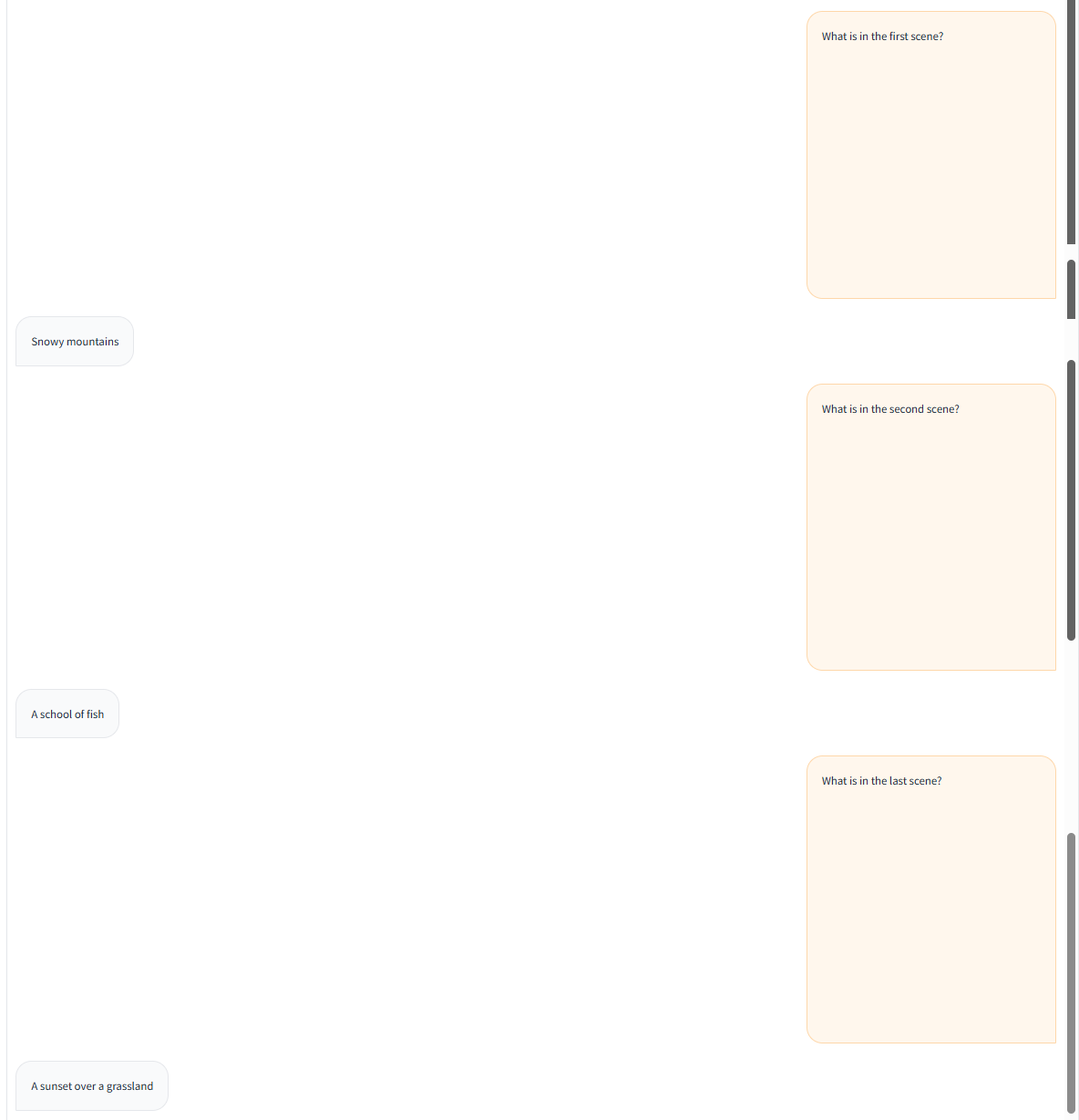}
    \vspace{-7pt}
    \caption{A video generated by \textit{Sora2} with the prompt: Present a video consisting of four scenes: snow mountain, underwater, desert, and grassland. Each scene should be highly realistic and appear to be filmed with a handheld camera.}
    \label{fig:case5}   
\end{figure*}

\begin{figure*}[t]
    \centering
    \includegraphics[width=0.9\linewidth]{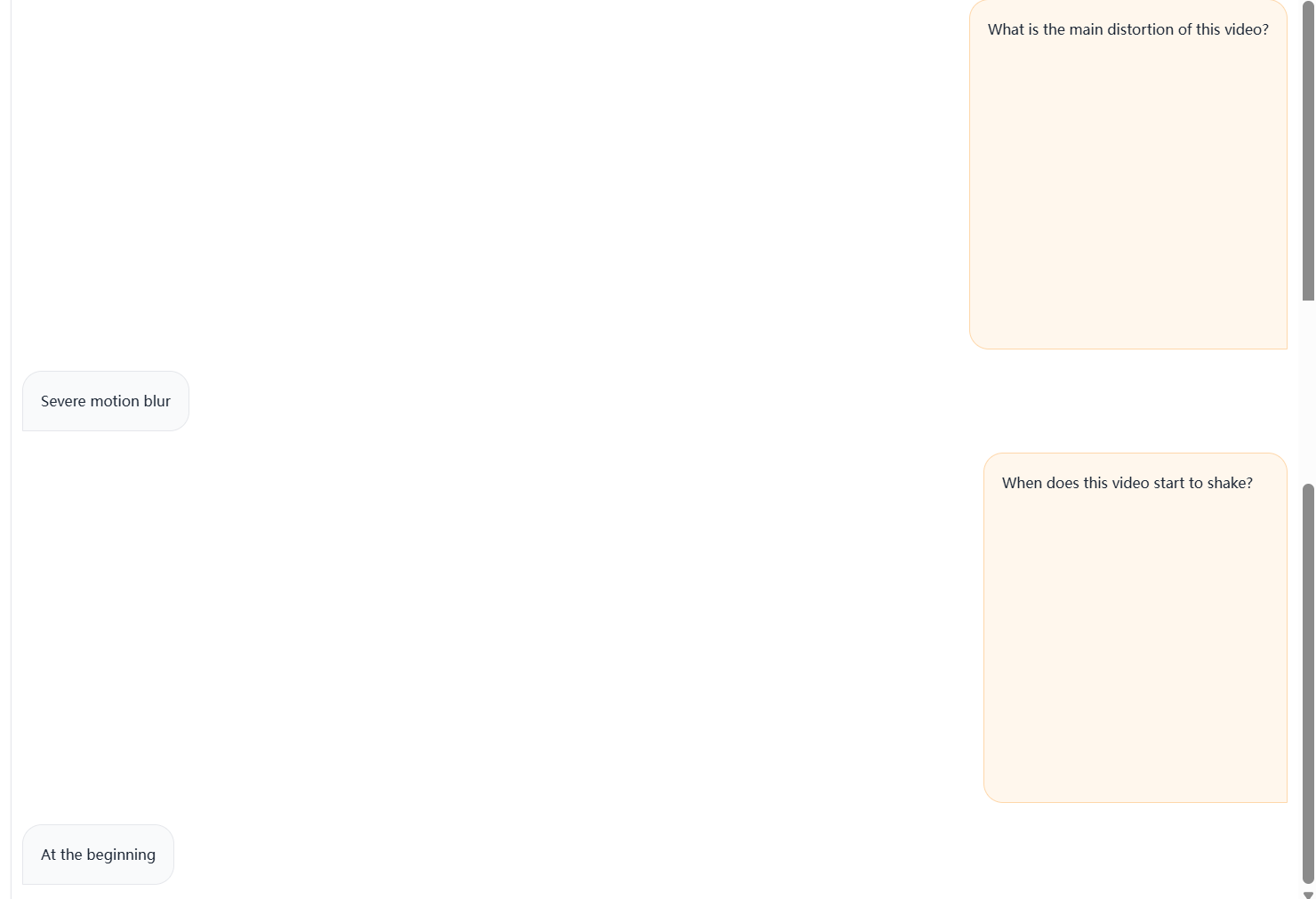}
    \vspace{-7pt}
    \caption{A video captured with a shaking camera in a staircase.}
    \label{fig:case6}   
\end{figure*}

Here we use \textit{Gradio Demo} and conduct six case studies with various image and video contents and types (shown in Figs.~\ref{fig:case1}, \ref{fig:case2}, \ref{fig:case3}, \ref{fig:case4}, \ref{fig:case5}, and \ref{fig:case6}).
\label{Case Studies}

\begin{figure*}[t]
    \centering
    \includegraphics[width=0.9\linewidth]{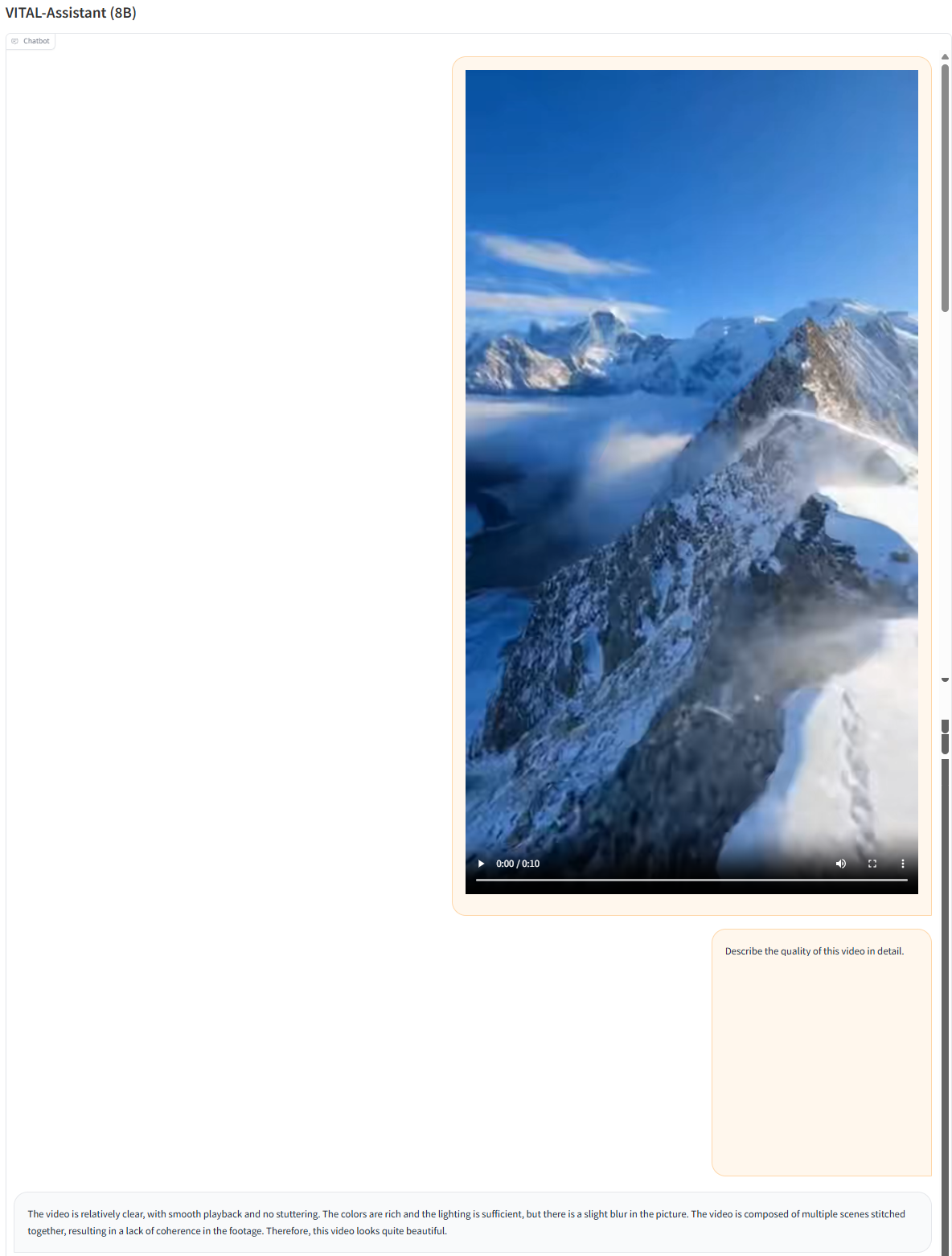}
    \vspace{-7pt}
\end{figure*}

\begin{figure*}[t]
    \centering
    \includegraphics[width=0.9\linewidth]{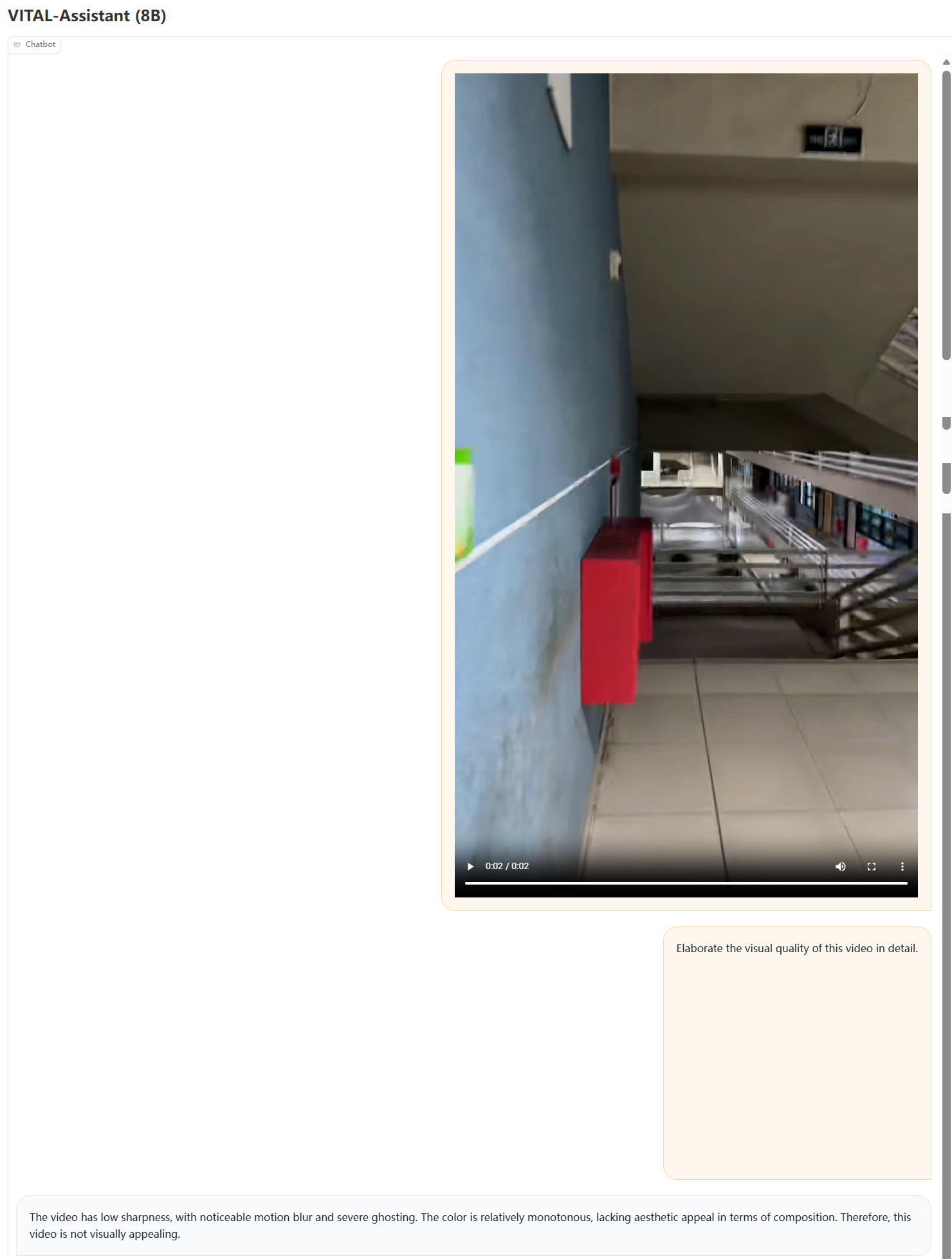}
    \vspace{-7pt} 
\end{figure*}

